\pdfoutput=1

\documentclass[11pt]{article}

\usepackage{acl}

\usepackage{times}
\usepackage{latexsym}
\usepackage[T1]{fontenc}
\usepackage[utf8]{inputenc}
\usepackage{microtype}
\usepackage{inconsolata}
\usepackage{booktabs, arydshln}
\usepackage{caption}
\usepackage{subcaption}
\usepackage{graphicx}
\usepackage{amssymb, bm}
\usepackage{tcolorbox}
\usepackage{comment}
\usepackage{color, colortbl}
\usepackage{siunitx}
\usepackage{multirow}
\usepackage{cleveref}

\definecolor{OurRed}{RGB}{240, 90, 50}
\definecolor{OurBlue}{RGB}{60, 70, 220}
\definecolor{OurCyan}{RGB}{40, 215, 250}

\makeatletter
\def\adl@drawiv#1#2#3{%
        \hskip.5\tabcolsep
        \xleaders#3{#2.5\@tempdimb #1{1}#2.5\@tempdimb}%
                #2\z@ plus1fil minus1fil\relax
        \hskip.5\tabcolsep}
\newcommand{\cdashlinelr}[1]{%
  \noalign{\vskip\aboverulesep
           \global\let\@dashdrawstore\adl@draw
           \global\let\adl@draw\adl@drawiv}
  \cdashline{#1}
  \noalign{\global\let\adl@draw\@dashdrawstore
           \vskip\belowrulesep}}
\makeatother

\title{The Impact of Demonstrations on Multilingual In-Context Learning:\\ A Multidimensional Analysis}

\author{Miaoran Zhang\textsuperscript{1} \,
 Vagrant Gautam\textsuperscript{1} \,
 Mingyang Wang\textsuperscript{2,3,4} \, 
 Jesujoba O. Alabi\textsuperscript{1} \vspace{3px} \  \\ 
 \textbf{Xiaoyu Shen\textsuperscript{5\thanks{\hspace{1mm} Corresponding author.}} }\, 
 \textbf{Dietrich Klakow\textsuperscript{1}} \, 
 \textbf{Marius Mosbach\textsuperscript{6}} \vspace{3px} \ \\ %
\textsuperscript{1}Saarland University, Saarland Informatic Campus \\
\textsuperscript{2}Bosch Center for AI
\textsuperscript{3}LMU Munich
\textsuperscript{4}Munich Center for Machine Learning (MCML)\\
\textsuperscript{5}Eastern Institute of Technology, Ningbo \ \
\textsuperscript{6}Mila, McGill University\\
{\tt  \{mzhang,vgautam,jalabi,dietrich.klakow\}@lsv.uni-saarland.de} \\ {\tt mingyang@cis.lmu.de\quad \tt xyshen@eitech.edu.cn \quad \tt marius.mosbach@mila.quebec}
}

\begin{document}
\maketitle

\begin{abstract}
In-context learning is a popular inference strategy where large language models solve a task using only a few labeled demonstrations without needing any parameter updates. Although there have been extensive studies on English in-context learning, multilingual in-context learning remains under-explored, and we lack an in-depth understanding of the role of demonstrations in this context. To address this gap, we conduct a multidimensional analysis of multilingual in-context learning, experimenting with 5 models from different model families, 9 datasets covering classification and generation tasks, and 56 typologically diverse languages. Our results reveal that the effectiveness of demonstrations varies significantly across models, tasks, and languages. We also find that strong instruction-following models including Llama 2-Chat, GPT-3.5, and GPT-4 are largely insensitive to the quality of demonstrations. Instead, a carefully crafted template often eliminates the benefits of demonstrations for some tasks and languages altogether.
These findings show that the importance of demonstrations might be overestimated.
Our work highlights the need for granular evaluation across multiple axes towards a better understanding of in-context learning.\footnote{We release our code publicly at \url{https://github.com/uds-lsv/multilingual-icl-analysis}.}

\end{abstract}

\section{Introduction}
\label{sec:intro}

An intriguing property of large language models (LLMs) is their ability to perform in-context learning~\citep{brown2020language}, i.e., solve a task conditioned on a few demonstrations at inference time, without updating the model parameters. It has been shown to be an efficient alternative to fine-tuning when adapting models to diverse tasks and domains~\citep[\textit{inter alia}]{dong2022survey, min-etal-2022-metaicl, si2023prompting}. In light of the success of in-context learning, there has been increased interest in better understanding the factors that influence its success, such as demonstration selection~\citep{liu-etal-2022-makes, rubin-etal-2022-learning, wang2023large}, prompt design~\citep{min-etal-2022-noisy, wei2022chain}, and more generally on understanding how and why in-context learning works~\citep{xie2022an, bansal-etal-2023-rethinking, hendel-etal-2023-context, pan-etal-2023-context, wang-etal-2023-label}.

However, most recent work on in-context learning predominantly focuses on English, and the exploration of multilingual in-context learning generally lags behind. This is problematic, as results that apply to English might not hold for other languages, especially those that are less represented in LLM training data. While there have been a few studies on in-context learning that go beyond English, they either focus on benchmarking LLMs on multilingual tasks without in-depth exploration, e.g., MEGA~\citep{ahuja-etal-2023-mega} and BUFFET~\citep{asai2023buffet}, or zoom in on specific capabilities such as mathematical reasoning~\citep{shi2023language},  machine translation~\citep{zhu2023multilingual, agrawal-etal-2023-context}, or code-switching~\citep{zhang-etal-2023-multilingual}.

In this work, we take a multidimensional approach~\citep{ruder-etal-2022-square} that unifies these strands of research and comprehensively evaluate the multilingual in-context learning abilities of LLMs. We focus on dissecting the \textit{actual} impact of in-context demonstrations, which is crucial for understanding model behaviour. Our research covers various models, tasks, and languages, and we seek to answer the following research questions: 

\vspace{-1mm}
\begin{enumerate}
    \item Does multilingual performance benefit from demonstrations? (\S\ref{sec:n_demo})
    \vspace{-2mm}
    \item Does demonstration quality matter? (\S\ref{sec:quality}) 
    \vspace{-2mm}
    \item What is the interplay between demonstrations and templates? (\S\ref{sec:prompt}) \vspace{-2mm}
    \item How do the answers to these questions vary across languages and models? (\S\ref{sec:n_demo}, \S\ref{sec:quality}, \S\ref{sec:prompt})
    \vspace{-1mm}
\end{enumerate}

\begin{figure*}[th!]
    \centering
    \includegraphics[width=\textwidth]{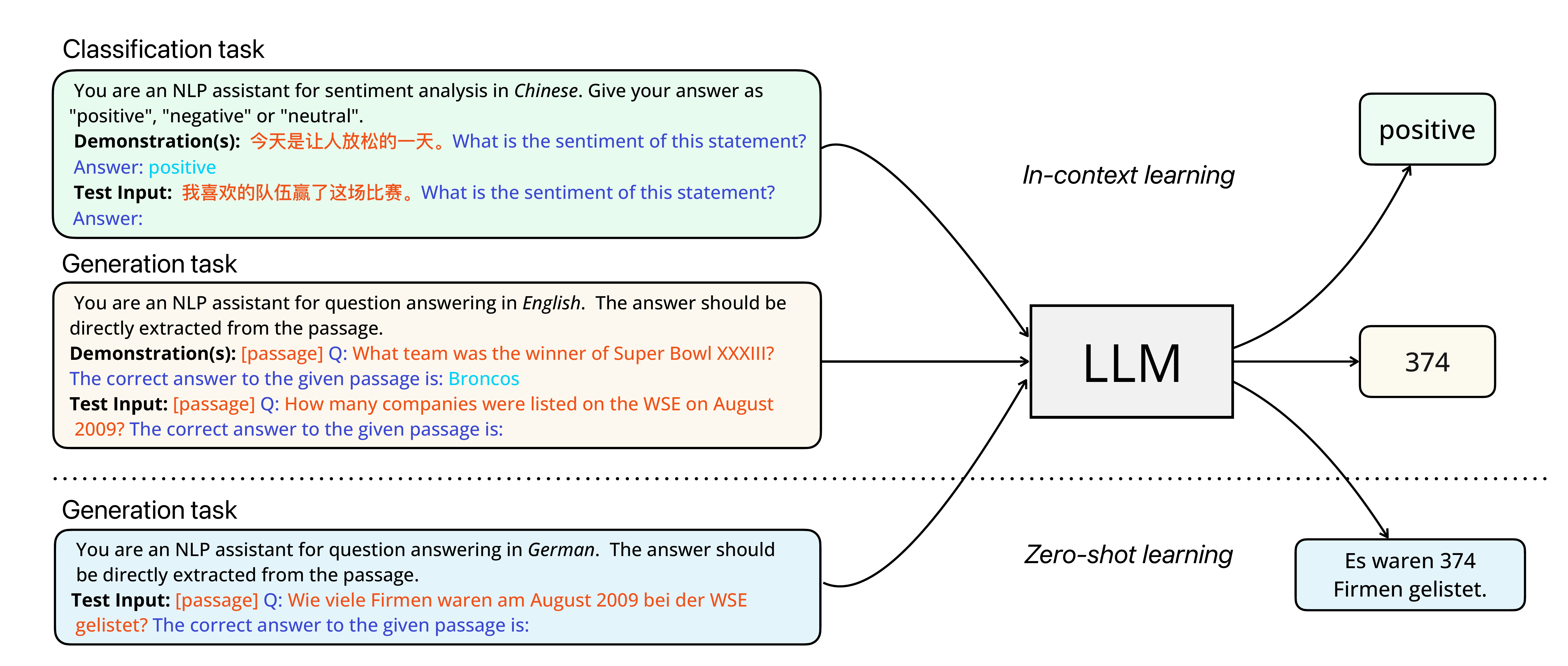}
    \vspace{-2em}
    \caption{An overview of the components of multilingual in-context learning (\S\ref{sec:in-context-learning}) with a comparison to zero-shot learning. Sources of variation include tasks, languages, models, and the template, i.e., the task instruction, \textcolor{OurBlue}{patterns} for formatting \textcolor{OurRed}{inputs}, and \textcolor{OurCyan}{verbalized labels}.}
    \label{fig:overview}
    \vspace{-2mm}
\end{figure*}
\noindent Specifically, we address our research questions by evaluating 5 LLMs including base models that are only pre-trained on unlabeled text corpora (XGLM and Llama 2), and chat models that are further refined with instruction tuning and reinforcement learning (Llama 2-Chat, GPT-3.5, and GPT-4). We evaluate on 9 multilingual datasets that include both classification and generation tasks, covering 56 typologically different languages. 

\raggedbottom
Our main findings are: 
(1) The effectiveness of demonstrations varies widely depending on the model, task, and language used. For base models, in-context learning barely outperforms zero-shot learning on many tasks. In general, in-context learning matters more for generation tasks with loosely-specified prompts; (2) Even with sophisticated demonstration selection methods, in-context learning is not always beneficial and can sometimes be worse than using no demonstrations at all; (3) Chat models are less sensitive to seeing correctly-labeled demonstrations than base models, suggesting that for the former, demonstrations primarily help the model understand the task format, while for the latter, demonstrations also impart task-specific knowledge; (4) Using a formatting-focused template can even eliminate the need for demonstrations with chat models. The relative significance of demonstrations versus prompt templates varies based on inherent model capabilities.

In sum, we suggest that the benefits of adding demonstrations may be overestimated. Future work on in-context learning should carefully compare their results with zero-shot learning and on multiple templates to faithfully represent its effectiveness. Given the vast variance across models, tasks, and languages, it is also important to cautiously frame claims about in-context learning.

\section{Preliminaries}
\label{sec:in-context-learning}
\subsection{In-context learning}
In-context learning (ICL) is a popular inference strategy where models solve\footnote{The extent to which models actually ``solve'' tasks is an open question as ICL, similar to fine-tuning, has generalization issues despite its impressive results~\citep{mosbach-etal-2023-shot}. Regardless, we use the word ``solve'' in the rest of this paper for simplicity.} a task without any parameter updates~\citep{brown2020language}. Instead, the model performs the task by conditioning on \textbf{labeled demonstrations}. Demonstrations are typically formatted using ``pattern-verbalizer pairs,'' as this has been shown to be effective in eliciting good task performance~\citep{schick-schutze-2021-just,bach-etal-2022-promptsource}. Here, a \textit{pattern} is used to format the input for the model, and a \textit{verbalizer} maps the label to a textual representation. Additionally for instruction-tuned LLMs, a \textit{task instruction} is often added to provide information about the task beyond individual demonstrations~\citep{ mishra-etal-2022-cross, wang-etal-2022-super, ouyang2022training}. 

Formally, given a test sample $x_t$, $k$ demonstrations $\{(x_i, y_i)\}_{i=1}^k$, a pattern $\mathcal{P}$, a verbalizer $\mathcal{V}$ and a task instruction $\mathcal{I}$, the model (parameterized by $\theta$) makes its prediction as follows:
\begin{equation}
    y_t \sim p_\theta(y | \mathcal{I}, \{(\mathcal{P}(x_i), \mathcal{V}(y_i))\}_{i=1}^k, \mathcal{P}(x_t)).
\end{equation}

\noindent Taken together, the pattern, the verbalizer and the optional task instruction comprise the \textbf{template} with which demonstrations and the test sample are formatted as the input prompt for model inference. The effectiveness of demonstrations is thus linked with the template used to present them to the model.

\subsection{Multilingual prompting}
\label{subsec:multilingual_prompting}

Previous studies highlight that the selection of demonstrations and prompt templates can significantly influence model performance~\citep{liu-etal-2022-makes, fu2023complexitybased, sclar2024quantifying}. In multilingual in-context learning, the variation in input prompts is further complicated by the \textit{language} of demonstrations, templates and test samples, all of which are important design choices.

For the template language, \citet{lin-etal-2022-shot} and \citet{ahuja-etal-2023-mega} found that English templates generally perform better than native language templates, possibly due to superior instruction-following abilities on existing LLMs on English compared to other languages.
Following this, we use English templates in our study.

For the language of few-shot demonstrations and test samples, there are three popular settings. Given a test sample in a certain language, the most straightforward approach is to use demonstrations in the same language (referred to as \textit{in-language demonstrations}). This setting directly measures the model's inherent ability to solve problems in that language. Another choice is to use \textit{English demonstrations} regardless of the language of the test sample. This is a cross-lingual transfer setup, where the goal is to transfer knowledge from a pivot language to a target language via in-context learning.
As highlighted in \citet{shi2023language} and \citet{ahuja-etal-2023-mega}, in-language demonstrations often outperform English demonstrations on diverse multilingual tasks.
Yet another option is to translate the test sample into English -- an approach called \textit{translate-test}, where the demonstrations are also in English.
While translate-test leads to strong performance~\citep{ahuja-etal-2023-mega}, this approach heavily relies on a translation system for data processing and centers the English proficiency of LLMs.
In this work, we are interested in dissecting the intrinsic multilingual capabilities of LLMs, therefore we choose to use \textbf{in-language demonstrations}.

All these design choices are represented visually in Figure~\ref{fig:overview}, which gives an overview of multilingual in-context learning. Detailed setup information is provided in the next section.

\raggedbottom

\begin{figure*}[t]
    \centering
    \includegraphics[width=\linewidth]{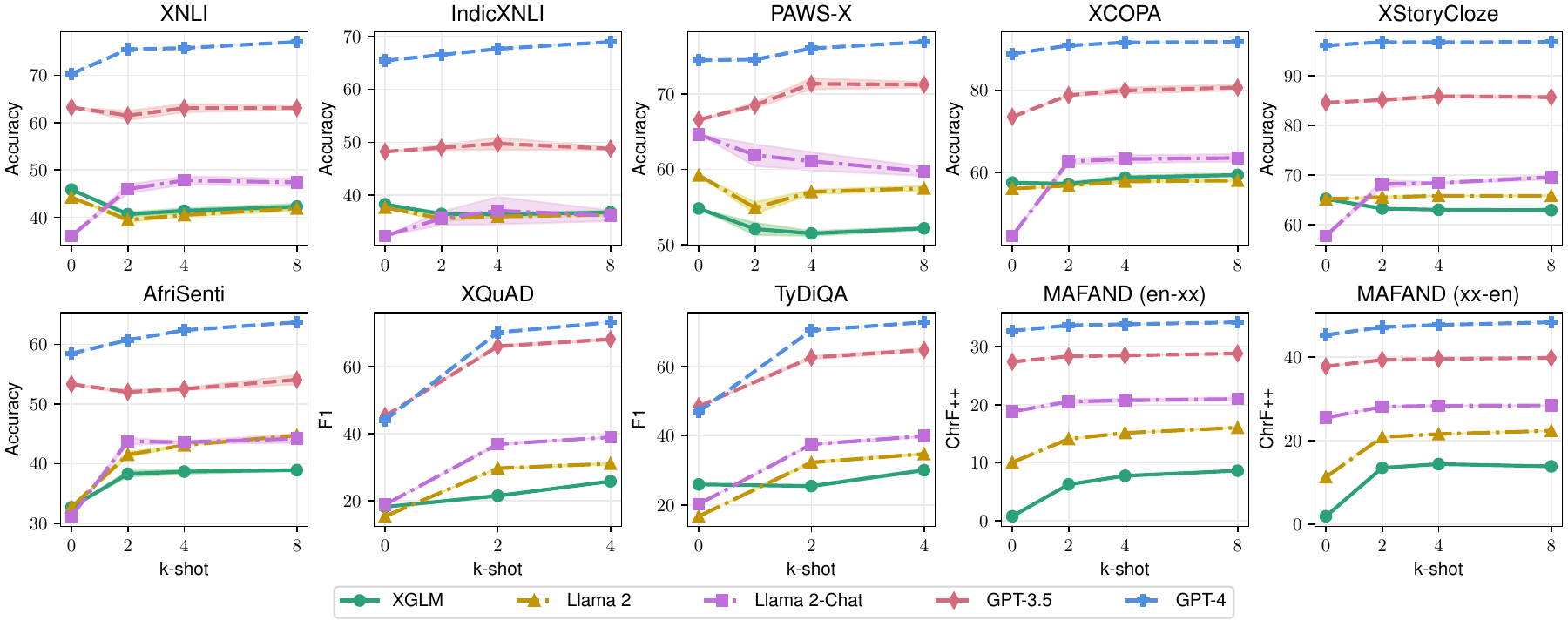}
    \vspace{-2em}
    \caption{Average performance across languages with different numbers of demonstrations. We average and report standard deviations over 3 seeds for all models except GPT-4. Note that the standard deviations are relatively small, possibly because of averaging over languages. en-xx: translating from English to another language, xx-en: translating from another language to English.}
    \label{fig:n_shot}
\end{figure*}

\section{Experimental setup}
\label{sec:setup}
\paragraph{Models.} We evaluate two types of LLMs: pre-trained base models and chat models. Our base models include XGLM~\citep{lin-etal-2022-shot} and Llama 2~\citep{touvron2023llama}. Our chat models are Llama 2-Chat, GPT-3.5~\citep{ouyang2022training} and GPT-4~\citep{openai2023gpt4}. Specifically, we use \texttt{xglm-7.5B}, \texttt{Llama-2-13b}, and \texttt{Llama-2-13b-chat} on Huggingface~\citep{wolf-etal-2020-transformers}, and we access \texttt{gpt-3.5-turbo-16k} and \texttt{gpt-4-32k} APIs via Microsoft Azure.\footnote{We also experiment with BLOOMZ and mT0~\citep{muennighoff-etal-2023-crosslingual}. Results in Appendix~\ref{sec:appendix_bloomz_mt0} show that their zero-shot performance significantly surpasses few-shot performance, which we ascribe to their training scheme.}

\raggedbottom
\paragraph{Tasks and datasets.} We experiment on a diverse range of multilingual classification and generation tasks, using 9 datasets covering 56 languages in total. Our dataset selection  largely follows MEGA~\citep{ahuja-etal-2023-mega}, but we add datasets for extremely under-represented African languages. Our classification tasks include natural language inference (NLI), paraphrase identification, commonsense reasoning and sentiment analysis, with the following datasets: XNLI~\citep{conneau-etal-2018-xnli}, IndicXNLI~\citep{aggarwal-etal-2022-indicxnli}, PAWS-X~\citep{yang-etal-2019-paws}, XCOPA~\citep{ponti-etal-2020-xcopa}, XStoryCloze~\citep{lin-etal-2022-shot} and AfriSenti~\citep{muhammad-etal-2023-afrisenti}. Our generation tasks are extractive question answering (QA) and machine translation (MT), for which we use XQuAD~\citep{artetxe-etal-2020-cross}, TyDiQA-GoldP~\citep{clark-etal-2020-tydi}, and MAFAND~\citep{adelani-etal-2022-thousand}. See Appendix~\ref{sec:appendix_tasks_datasets} for more details.

\paragraph{In-context learning.} For each test sample, we select $k\in\{0, 2, 4, 8\}$\footnote{For QA datasets, we select a maximum of $4$ demonstrations due to context size limitations.} different demonstrations, which are randomly sampled unless otherwise specified. All demonstrations are in the same language as the test sample, and all templates are in English. We employ appropriate task-specific templates for different model types. All templates and data splits are shown in Appendix~\ref{sec:appendix_prompting_setup}.

\paragraph{Metrics.} For classification tasks, we report the rank classification accuracy\footnote{The scoring function is the average of per-token log probabilities (ignoring the common prefix of different candidates). The candidate with the highest score is chosen as the prediction.} for open-source base models~\citep{muennighoff-etal-2023-crosslingual, lin-etal-2022-shot}. For chat models, we measure the exact match between generated outputs\footnote{We extract verbalized labels from the generated outputs using regular expressions before calculating the exact match.} and verbalized labels~\citep{ahuja-etal-2023-mega}. As for generation tasks, we use the F1 score for QA datasets and ChrF++ score~\citep{popovic-2017-chrf} for MAFAND. Implementation details for our evaluation are provided in Appendix~\ref{sec:appendix_implementation}.

\section{Do (more) demonstrations benefit multilingual performance?}
\label{sec:n_demo}

In this section, we systematically compare ICL and zero-shot learning as this question is under-explored in previous studies of multilingual ICL~\citep{ahuja-etal-2023-mega, asai2023buffet}. We examine model performance on diverse multilingual tasks while varying the number of demonstrations, and show the results for classification tasks and generation tasks in Figure~\ref{fig:n_shot}.

We begin with the overall trends across models and datasets. OpenAI's GPT-3.5 and GPT-4 models achieve the best multilingual in-context learning performance on all our datasets, which is unsurprising as they are currently the state-of-the-art on a large suite of NLP benchmarks.
The next best models are Llama 2 and Llama 2-Chat, which demonstrate comparable or superior performance to the multilingual XGLM model despite being trained primarily on English corpora \citep{touvron2023llama}. This indicates that their task-solving abilities can transfer across languages.
Regardless of the model, however, performance on the AfriSenti and MAFAND datasets, particularly when translating English to African languages, lags significantly behind other tasks, showing that language discrepancies remain even in the best models.\looseness-1

\begin{figure}[t]
    \centering
    \includegraphics[width=\columnwidth]{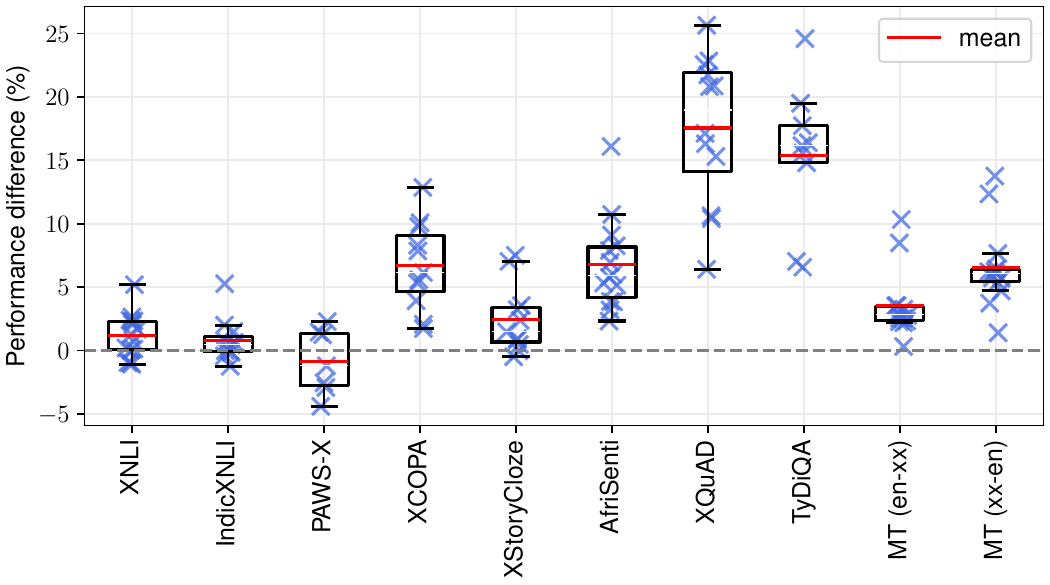}
    \caption{Performance difference between 4-shot and 0-shot. Each marker represents the average performance across models for each language in a given task. MT denotes the MAFAND dataset.}
    \label{fig:task_lang_dist}
\end{figure}

\definecolor{Color}{RGB}{250, 210, 210}
\definecolor{ColorText}{RGB}{220, 90, 90}

\begin{table*}[t]
  \centering
  \scalebox{0.7}{
    \begin{tabular}{l| S[table-format=1.2] S[table-format=1.2] S[table-format=1.2] S[table-format=1.2] S[table-format=1.2] S[table-format=1.2] | S[table-format=2.2] S[table-format=2.2] S[table-format=1.2] S[table-format=1.2]}
      \toprule
       \textbf{Model} &  \textbf{XNLI} & \textbf{IndicXNLI} & \textbf{PAWS-X} & \textbf{XCOPA} & \textbf{XStoryCloze} & \textbf{AfriSenti} & \textbf{XQuAD} & \textbf{TyDiQA}  & \textbf{MT (en-xx)} & \textbf{MT (xx-en)}  \\ 
        \midrule
         XGLM     &   4.59 &        2.49 &     0.24$_\triangledown$ &    0.03 &          0.97$_\triangledown$ &        5.62 &    1.77 &     4.21 &             1.31 &             0.66 \\
         Llama 2      &   6.61 &        4.17 &     2.35 &   \cellcolor{Color} -0.11 &          0.33 &        4.17 &    1.32 &     0.54 &             2.15 &             1.35 \\
         \cdashlinelr{1-11}
         Llama 2-Chat &  \cellcolor{Color}-0.28 &    \cellcolor{Color} -1.36 &  \cellcolor{Color}   -1.71$_\triangledown$ &    0.32 &          0.43 &        2.17 &    1.02 &     2.42 &             0.74 &             0.66 \\
         GPT-3.5      &   0.18 &        0.71 &   \cellcolor{Color}  -2.07 &    0.86 &        \cellcolor{Color}  -0.61 &        \cellcolor{Color}-0.66$_\triangledown$ &    \cellcolor{Color}-0.34 &     2.98 &             0.72 &             0.43 \\
         GPT-4        &   0.76 &     \cellcolor{Color}   -0.19 &     0.07 &   \cellcolor{Color} -0.36 &          0.05 &     \cellcolor{Color}   -0.68 &  \cellcolor{Color}  -0.77 &     1.88 &             1.21 &             0.65 \\
      \bottomrule
    \end{tabular}}
  \caption{Performance difference of 4-shot ICL with \textsc{top-k} vs. \textsc{random} selection. Positive numbers show that \textsc{top-k} is better than \textsc{random} (expected), and  highlighted cells show where \textcolor{ColorText}{top-k is even worse than random selection}. $\triangledown$: \textsc{top-k} performance is even worse than zero-shot learning. For \textsc{random}, we average over 3 seeds (except for GPT-4).} %
  \label{tab:demo_diff_topk}
\end{table*}

\definecolor{Color}{RGB}{190, 210, 240}
\definecolor{ColorText}{RGB}{60, 100, 210}

\begin{table*}[t]
  \centering
  \scalebox{0.7}{
    \begin{tabular}{l| S[table-format=1.2] S[table-format=1.2] S[table-format=1.2] S[table-format=1.2] S[table-format=1.2] S[table-format=1.2] | S[table-format=2.2] S[table-format=2.2] S[table-format=1.2] S[table-format=1.2]}
      \toprule
       \textbf{Model} &  \textbf{XNLI} & \textbf{IndicXNLI} & \textbf{PAWS-X} & \textbf{XCOPA} & \textbf{XStoryCloze} & \textbf{AfriSenti} & \textbf{XQuAD} & \textbf{TyDiQA}  & \textbf{MT (en-xx)} & \textbf{MT (xx-en)}  \\ 
       \midrule
       XGLM         & 0.46     &   \cellcolor{Color}    -0.05 &     0.44 & 0.51     & 0.62$^*$      & 3.78$^*$    & 24.56$^*$ & 26.64$^*$ & 3.18$^*$     & 6.73$^*$     \\
       Llama 2      & 0.96$^*$ &        0.43 &     1.16 & 0.61$^*$ & 1.12$^*$      & 2.27$^*$    & 26.68$^*$ & 29.20$^*$  & 4.79$^*$     & 8.34$^*$     \\
       \cdashlinelr{1-11}
       Llama 2-Chat & \cellcolor{Color} -0.34    &        0.04 &     1.48 & 0.03     & \cellcolor{Color} -0.23         & 0.77$^*$    & 5.94$^*$  & 4.37$^*$  & 1.13$^*$     & 1.53$^*$     \\
       GPT-3.5      & 0.39     &        1.02 &     0.64 & 0.26     & 0.58$^*$      & \cellcolor{Color} -0.62       & 5.46$^*$  & 5.61$^*$  & 1.39$^*$     & 0.48$^*$     \\
       GPT-4        & \cellcolor{Color} -0.86    &   \cellcolor{Color}    -0.04 &     0.57 & 0.86     & 1.13          & 0.90        & 9.60       & 6.97      & 1.24         & 0.64         \\
      \bottomrule
    \end{tabular}}
  \caption{Performance difference of 4-shot ICL with \textsc{random} vs. \textsc{random-corrupted} demonstrations. Positive numbers show that \textsc{random} is better than \textsc{random-corrupted} (expected), and highlighted cells show where \textcolor{ColorText}{corrupted labels perform even better than ground-truth labels}. We average over 3 seeds (except for GPT-4). $*$: a significant difference ($p=0.05$).}
  \label{tab:demo_diff_corrupt}
\end{table*}

An important pattern across datasets and models is that \textbf{in-context learning does not always improve over zero-shot learning} -- in particular, it helps with generation tasks, but results on classification tasks are mixed. For the AfriSenti dataset, many models show noticeable improvements with ICL. However, with other tasks such as IndicXNLI, XNLI and PAWS-X, the same models, especially base models, perform much worse compared to the zero-shot setting. We also see marginal improvements in some cases, e.g., XGLM and Llama 2 on XCOPA. In comparison to chat models, the addition of demonstrations typically reduces the performance of base models across many tasks. When examining the cases where ICL improves performance, we see that \textbf{improvements saturate quickly with 2 to 4 demonstrations}. This aligns with \citet{chen-etal-2023-many}, who found that reducing the number of demonstrations to one does not significantly deteriorate chain-of-thought reasoning.

\begin{figure}[t]
    \centering
    \includegraphics[width=\columnwidth]{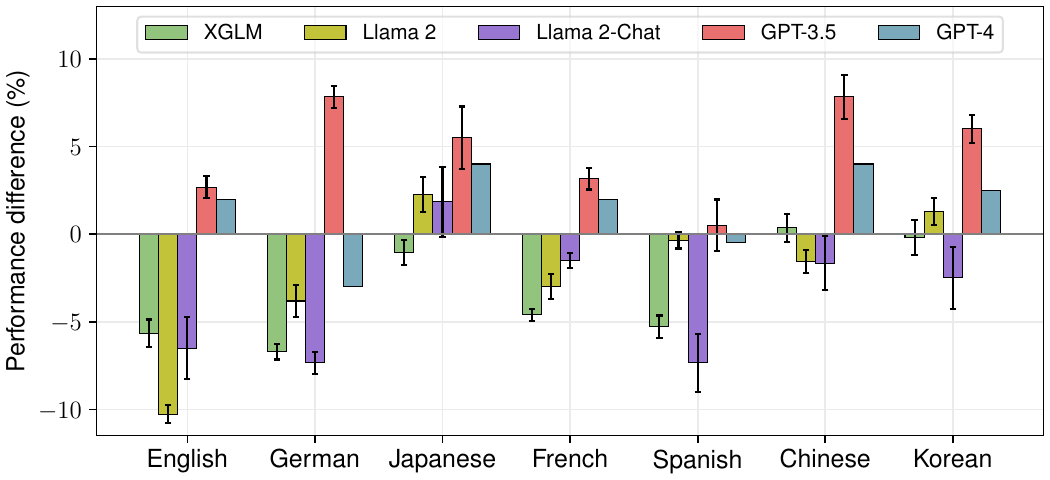}
    \caption{Performance difference between 4-shot and 0-shot for individual languages in PAWS-X. Error bars represent standard deviations calculated over 3 seeds.}
    \label{fig:pawsx_lang_dist}
\end{figure}

Looking at the improvements over zero-shot performance (for all models and languages combined) across tasks in Figure~\ref{fig:task_lang_dist}, we observe that there are large fluctuations between individual languages that are not captured by the average. The PAWS-X dataset in particular shows an average degradation, but in fact some languages benefit from ICL while others degrade. For a more nuanced understanding of language-specific differences within a task, we zoom into this dataset in Figure~\ref{fig:pawsx_lang_dist} to inspect these language-specific differences.\footnote{Plots for other datasets are provided in Appendix~\ref{sec:appendix_n_shot_languages}.} We see that languages and models can behave very differently even on just one dataset, and a pattern which holds for one language with one model does not necessarily apply to a different language. For example, the ICL performance of Llama~2 outperforms its zero-shot performance by 2.3 points on Japanese and 1.3 points on Korean. However, demonstrations degrade performance for other languages, e.g., English performance degrades by 10.3 points. In sum, \textbf{the effectiveness of demonstrations varies widely depending on the model, task, and language.}

\section{Does demonstration quality matter?}
\label{sec:quality}

\begin{figure*}[t] 
    \centering
    \includegraphics[width=\linewidth]{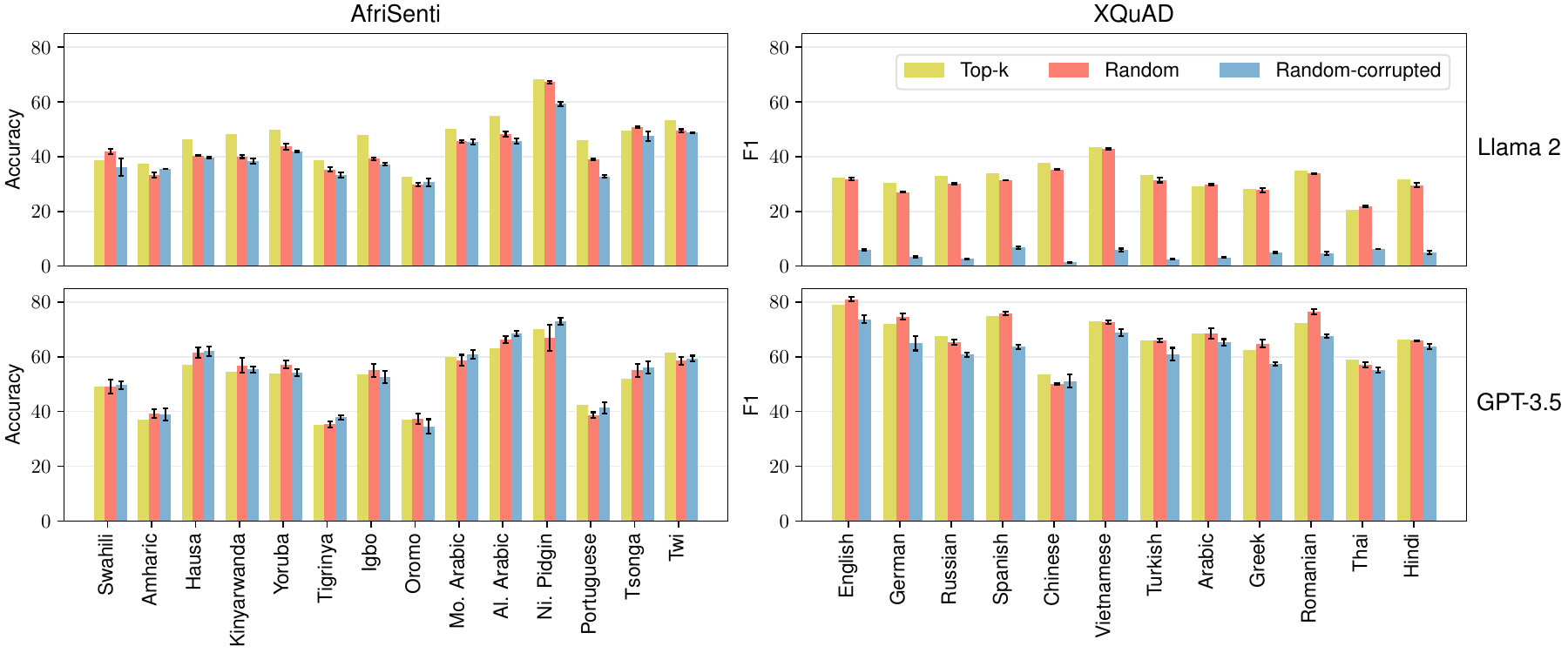}
    \vspace{-2em}
    \caption{Performance of 4-shot ICL using different types of demonstrations for individual languages on AfriSenti and XQuAD. The top row shows Llama 2 results, and the bottom row shows GPT-3.5 results.}
    \label{fig:task_lang_selection}
\end{figure*}

Our previous experiments evaluated ICL using randomly selected demonstrations. To ablate for the effects of demonstration quality, this section experiments with the choice of demonstrations as well as the importance of ground truth labels, i.e., the input-label mapping. Inspired by work on demonstration selection~\citep{liu-etal-2022-makes, rubin-etal-2022-learning} and input-label mapping~\citep{min-etal-2022-rethinking, yoo-etal-2022-ground} in English, we compare the following three types of demonstrations:

\begin{itemize}
    \item \textbf{\textsc{Random}}: demonstrations are randomly selected from clean data
    \item \textbf{\textsc{Top-k}}: the $k$ most semantically similar\footnote{We quantify semantic similarity using  LaBSE~\citep{feng-etal-2022-language}, a multilingual sentence embedding model trained on 109+ languages.} examples to a given test sample are selected~\citep{liu-etal-2022-makes}
    \item \textbf{\textsc{Random-corrupted}}: demonstrations are randomly selected but the labels are corrupted by replacement with random labels\footnote{For classification tasks, we randomly choose a label from the fixed label set. For generation tasks, we randomly choose a label from the label space of the entire demonstration data.}~\citep{min-etal-2022-rethinking}
\end{itemize}

Table~\ref{tab:demo_diff_topk} shows that top-k selection performs better than random selection in many cases, especially for the base models XGLM and Llama 2.
For chat models, the largest improvements are on generation tasks.
For example, GPT-3.5 achieves a 2.98-point improvement on TyDiQA. Nevertheless, top-k selection often degrades performance on many other tasks, e.g., GPT-3.5 is 2.07 points worse on PAWS-X compared to random selection. When compared to zero-shot performance, ICL with top-k selection is even \textit{worse} in some cases, such as XGLM on PAWS-X and XStoryCloze. In cases where random selection performs worse than zero-shot, even top-k selection gives only marginal improvements (see detailed numbers in Table~\ref{tab:demos} in Appendix~\ref{sec:appendix_demo_performance}). These findings indicate that \textbf{sophisticated demonstration selection methods are not always beneficial and can sometimes be worse than using no demonstrations at all}.

Exploring this further, in Table~\ref{tab:demo_diff_corrupt}, we compare randomly selected demonstrations with ground truth labels and corrupted labels. We find that using corrupted labels does not hurt performance on multilingual classification tasks much, which is consistent with previous research on English~\citep{min-etal-2022-rethinking}. On generation tasks, however, all models perform worse with corrupted labels, but to vastly different extents. XGLM and Llama 2 perform significantly worse with corrupted labels, especially on the machine translation task, whereas \textbf{chat models do not rely as much on correct labels}. This might be explained by ICL helping the model understand the task format and activating prior knowledge acquired by the model, rather than the model learning the task from demonstrations.
The observed model insensitivity to correct labels on certain tasks implies that random labels can serve as a strong baseline for demonstration generation before exploring more complex methods~\citep{lyu-etal-2023-z, wan-etal-2023-universal}. 

To investigate how these patterns split up across languages, Figure~\ref{fig:task_lang_selection} shows language-specific results on AfriSenti and XQuAD with Llama 2 and GPT-3.5.\footnote{See Appendix~\ref{sec:appendix_quality_languages} for other models and datasets.}
On AfriSenti, top-k selection outperforms random selection with Llama~2 across most languages; however, in the case of Swahili and Tsonga, there is a performance drop of 3.2 and 1.2 points, respectively. With GPT-3.5, top-k selection does not help across most languages, but it does help with Mozambican Portuguese and Twi. Similarly, the impact of corrupted labels varies. Llama 2 is affected dramatically by corrupted labels on all languages in XQuAD, whereas GPT-3.5 is much less affected, although to varying degrees across different languages. We urge NLP practitioners to \textbf{attend to these discrepancies when creating language-specific applications}, and leave it to future work to explore where they come from.

\section{Better templates further reduce the benefits of demonstrations}
\label{sec:prompt}

\noindent In-context learning performance depends not only on the demonstrations, which we have varied so far, but also on how they are formatted using templates. Previous work~\citep{gonen-etal-2023-demystifying, mizrahi2024state} has shown that modifying the template changes task performance. This section thus seeks to examine the interplay between template choice and demonstrations.

\paragraph{Template design.}
In the zero-shot setting, we observe that chat models tend to generate verbose responses (e.g., ``Sure! I can help you with that'') or explanations (e.g., ``The reason is that ...'') that pose a challenge for automatic evaluation. We observe a reduction in this behaviour with ICL, which leads us to question whether demonstrations are merely a means to format model responses. To see if we can achieve the same effect with minor template engineering, we augment the original templates with instructions that focus on output formatting. We call these \textit{formatting-focused templates} which are shown in Table~\ref{tab:template_openai_1}.

\begin{figure}[!htb]
    \centering
    \includegraphics[width=\columnwidth]{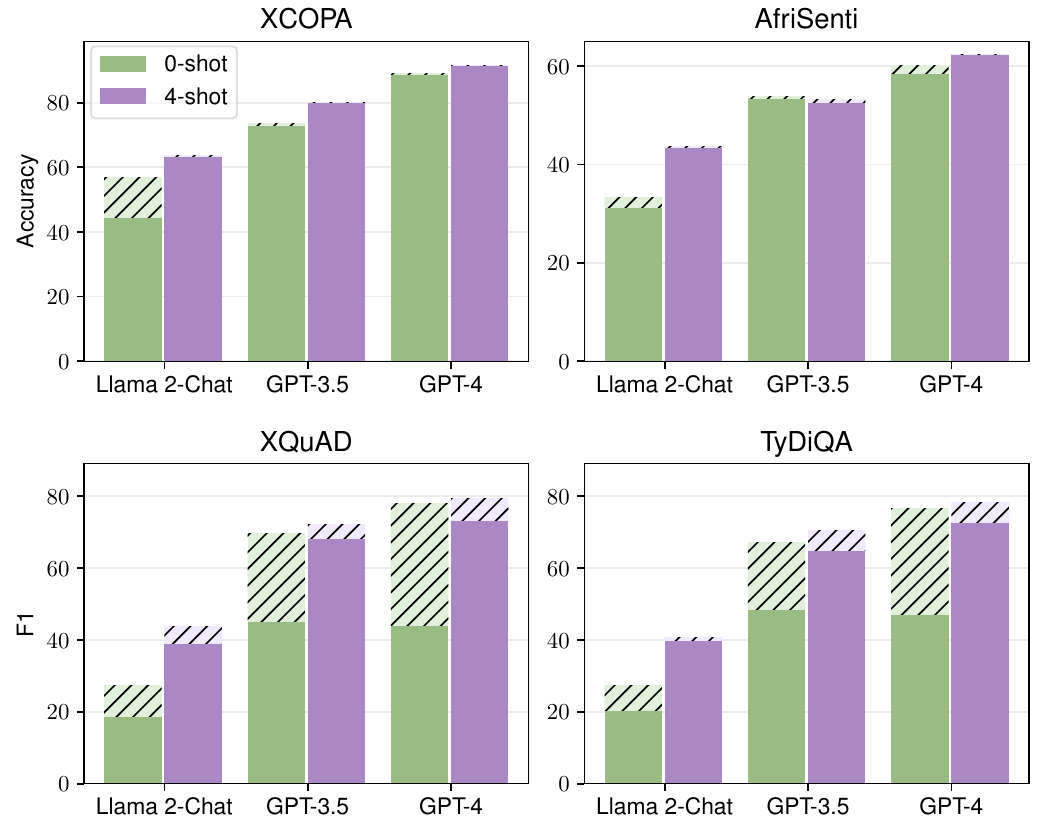}
    \caption{Effect of using different templates on 0-shot and 4-shot performance. Formatting-focused templates (with hatching) improve 0-shot performance over original templates (solid colours), and reduce the gap between 0-shot and 4-shot performance. Few-shot results are averaged across 3 seeds except for GPT-4.}
    \label{fig:template}
    \vspace{2mm}
\end{figure}

In this section, we focus on XCOPA, AfriSenti, XQuAD, and TyDiQA, as these are the classification and generation tasks that seem to benefit most from in-context demonstrations (see Section~\ref{sec:n_demo}). However, as Figure~\ref{fig:template} shows, \textbf{the performance gap between zero-shot and in-context learning diminishes with formatting-focused templates}. The gap reduction is more substantial for QA datasets (i.e., the generation tasks) than for XCOPA and AfriSenti (i.e., the classification tasks). We speculate that it is simpler for the model to generate label words for classification tasks with a pre-defined label space than to answer questions in a way that is easy to evaluate automatically. In the latter case, formatting-focused templates can teach output styling, largely eliminating the benefits of demonstrations.

Compared to GPT-3.5 and GPT-4, Llama 2-Chat performs worse in both zero-shot and few-shot settings, and formatting-focused templates have a less pronounced impact. On QA datasets, GPT-3.5 and GPT-4 even achieve better zero-shot performance with formatting-focused templates than ICL with original templates, a pattern that is not observed with Llama 2-Chat. This suggests that \textbf{the relative significance of demonstrations and templates varies based on the inherent abilities of models} at solving tasks and following instructions.

\definecolor{Color}{rgb}{0.9, 0.9, 0.9}

\begin{table}[t]
  \centering
  \scalebox{0.62}{
    \begin{tabular}{llrrrr}
      \toprule
      \multirow{2}{*}{\textbf{Model}} & \multirow{2}{*}{\textbf{Demo. Label}} &  \multicolumn{2}{c}{\textbf{XQuAD}} &  \multicolumn{2}{c}{\textbf{TyDiQA}} \\ 
      \cmidrule{3-6}
       && {O} & {F} & {O} & {F} \\
      \midrule
      & Original &  38.9$_{\pm0.1}$ &  43.8$_{\pm0.7}$ &  40.0$_{\pm0.3}$ &  40.6$_{\pm0.8}$  \\
      Llama 2-Chat& Corrupted &  33.0$_{\pm0.4}$ &  38.6$_{\pm0.3}$ & 35.6$_{\pm0.1}$  &  36.3$_{\pm0.5}$ \\
      & \cellcolor{Color}$\Delta$ &\cellcolor{Color}5.9 &\cellcolor{Color}5.2 &\cellcolor{Color}4.4& \cellcolor{Color}4.3 \\
       \midrule
      &Original &  68.2$_{\pm0.4}$ & 72.2$_{\pm0.4}$ & 64.8$_{\pm0.5}$ & 70.5$_{\pm0.5}$ \\
      GPT-3.5&Corrupted & 62.7$_{\pm0.2}$ &	69.9$_{\pm0.2}$ &	59.2$_{\pm0.3}$&	67.1$_{\pm0.7}$ \\
      &\cellcolor{Color} $\Delta$ & \cellcolor{Color}5.5	& \cellcolor{Color}2.3	& \cellcolor{Color}5.6	& \cellcolor{Color}3.4 \\
      \midrule
      &Original & 73.2 &	79.3 &	72.8 &	78.3\\
      GPT-4 &Corrupted & 63.6 &	79.8 &	65.8 &	77.6\\
      &\cellcolor{Color} $\Delta$ &\cellcolor{Color}9.6  &\cellcolor{Color}-0.5 &\cellcolor{Color}7.0 &\cellcolor{Color}0.7 \\
      \bottomrule
  \end{tabular}}
  \caption{Effect of using different templates on 4-shot performance with \textsc{random} and \textsc{random-corrupted} demonstrations. When using formatting-focused templates (F) over the original templates (O), the performance gap ($\Delta$) between original and corrupted labels decreases. We average and report standard deviations over 3 seeds for all models except GPT-4.}
  \label{tab:demo_diff_template}
\end{table}

With our new formatting-focused templates, we revisit the impact of the input-label mapping discussed in Section~\ref{sec:quality}. 
As ~\Cref{tab:demo_diff_template} shows, all models perform worse with corrupted labels, but formatting-focused templates largely mitigate this degradation. Notably, \textbf{GPT-4 using corrupted labels performs on par with ground truth labels}. This strengthens our finding that the correct input-label mapping is not that important, while also highlighting the crucial role that templates play in in-context learning.

\begin{figure}[t]
    \centering
    \includegraphics[width=\columnwidth]{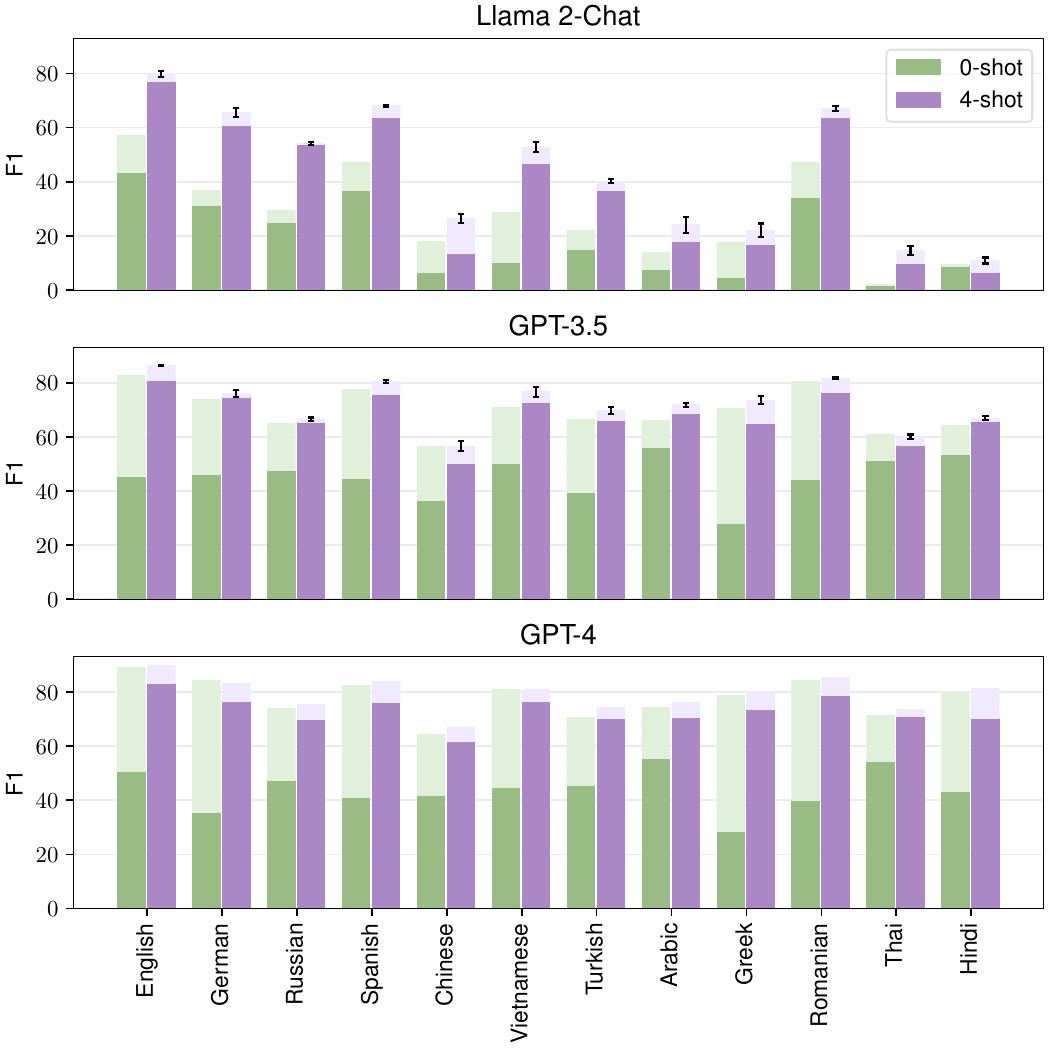}
    \caption{Effect of using different templates on 0-shot and 4-shot XQuAD performance. Formatting-focused templates (with hatching) improve 0-shot performance over original templates (solid colours), and reduce the gap between 0-shot and 4-shot performance. Few-shot results are averaged across 3 seeds except for GPT-4.}
    \label{fig:xquad}
    \vspace{-2mm}
\end{figure}

\Cref{fig:xquad} shows the language-specific effects of formatting-focused templates on XQuAD (results for other tasks are in Appendix~\ref{sec:appendix_template_languages}). For Llama 2-Chat, demonstrations remain essential even with a formatting-focused template for most languages, but not Greek and Hindi. GPT-3.5 and GPT-4 also show variance across languages. Moreover, for most languages, zero-shot learning with minor template engineering can match and even exceed in-context learning performance, aligning with previous work on GPT-3~\citep{reynolds2021prompt}. The fact that we can achieve the same effects through template engineering or demonstrations reinforces our hypothesis that models are not actually learning tasks on the fly. Instead, some combination of demonstrations and templates serves to activate prior knowledge of a task and encourage a consistent output format for automatic evaluation.

\section{Discussion}
\label{sec:discussion}

Our systematic study provides strong evidence that the importance of in-context demonstrations on existing multilingual datasets might be overestimated, as it highly depends on the model, task, and language used. For strong instruction-following models, the effect of demonstrations is \textit{superficial} and can be eliminated with minor template engineering. These findings open up new questions, which we discuss below.

\paragraph{Understanding the failures of ICL.} There has been a surge of research interest in understanding the underlying mechanisms of ICL~\citep{xie2022an, pmlr-v202-von-oswald23a, wang-etal-2023-label, hendel-etal-2023-context}, motivated by its successes. Our results show that ICL is \textit{not} always effective, and that its performance changes depending on multiple factors including the choice of model, task and language. The failures of ICL need as much scrutiny as its successes for a more fundamental understanding of the learning mechanisms of LLMs.

\paragraph{Optimizing demonstrations or templates.} With the increasing popularity of research on demonstration selection~\citep{liu-etal-2022-makes, rubin-etal-2022-learning, li-etal-2023-unified} and prompt engineering~\citep{mishra-etal-2022-reframing, white2023prompt, khattab2023dspy}, it is important to understand the interplay of the two. We show that good demonstrations help base models perform better on certain tasks, but that formatting-focused prompting has a much bigger impact on chat models. These results show that the impact of demonstrations cannot be fairly evaluated in isolation from the choice of prompt. These findings have implications both for researchers interested in fairly evaluating ICL, and for practitioners to choose to spend time optimizing demonstrations, templates or both.

\paragraph{Evaluating multilingual ICL.} 
Compared to the extensive research on ICL in English~\citep{zhao2021calibrate,dong2022survey,min-etal-2022-metaicl, mosbach-etal-2023-shot}, multilingual ICL remains under-explored. There is no widely accepted setup to robustly evaluate the effectiveness of ICL across languages, since the choice of multilingual models and tasks is limited. Based on our findings, we have some recommendations for the nascent field of multilingual ICL. First, critical evaluation is important. We need to compare ICL strategies to zero-shot learning, and ablate them with multiple templates. Second, as there is so much variance across models, tasks and languages, it is important to carefully scope claims about ICL. Last but not least, every language is different, so granular per-language analysis is a must in multilingual research.\looseness=-1

\section{Related Work}
\label{sec:related-work}
\paragraph{Multilingual in-context learning.}
Most multilingual in-context learning studies focus on benchmarking LLMs on diverse tasks and comparing them with smaller fine-tuned models~\citep{ahuja-etal-2023-mega, asai2023buffet, zhang-etal-2023-multilingual, zhu2023multilingual}.
As these works focus on benchmarking, their analysis of the role of demonstrations is limited.
\citet{ahuja-etal-2023-mega} explore different prompting strategies by adjusting the language of templates and demonstrations. \citet{zhang-etal-2023-multilingual} find that demonstrations sometimes do not contribute to or even degrade model performance on code-switching. \citet{zhu2023multilingual} look at machine translation and analyze the effects of template and demonstration selection with XGLM.
In the context of cross-lingual transfer, \citet{shi-etal-2022-xricl}, \citet{tanwar-etal-2023-multilingual}, and \citet{agrawal-etal-2023-context} investigate demonstration selection for specific applications. 
In contrast, we take a much broader perspective and investigate the actual impact of demonstrations across a wide range of models, tasks and languages.

\paragraph{English-centric demonstration analysis.} Most of the current demonstration analysis literature focuses on English: 
\citet{lu-etal-2022-fantastically} analyze the sensitivity of ICL to the order of demonstrations, \citet{min-etal-2022-rethinking} and \citet{yoo-etal-2022-ground} explore whether the ground truth labels matter for classification tasks, and \citet{wei2023larger} investigate the sensitivity of various model families to different input-label mappings. Similarly, \citet{pan-etal-2023-context} disentangle task recognition and task learning by manipulating the label space. Beyond this, \citet{shi2023large} and \citet{wang-etal-2023-towards} modify the validity of chain-of-thought (CoT) reasoning steps in demonstrations and explore the impact of this modification on mathematical reasoning. Also focusing on CoT, \citet{chen-etal-2023-many} investigate how varying the number of demonstrations affects performance.

\section{Conclusion}
\label{sec:conclusion}
In this paper, we conduct an in-depth multidimensional analysis on the impact of demonstrations in multilingual in-context learning. We find that the use of demonstrations does not always provide benefits compared to zero-shot learning, and that there is a large variance in performance across models, datasets and languages. While the quality of demonstrations influences the performance of base LLMs on certain tasks, the impact is significantly reduced for LLMs tuned with alignment techniques. We also examine the interplay between demonstrations and templates, finding that a carefully crafted template can further decrease the benefits of demonstrations. Our granular analysis contributes novel insights with nuance and paves the way for a more thoughtful multilingual ICL evaluation. 

\section*{Limitations}
\paragraph{Data contamination.} Since LLMs are trained with a vast amount of data scraped from the internet, this might result in data contamination, i.e., when the training data includes test datasets. \citet{ahuja-etal-2023-mega} suspect that many multilingual datasets appear in the training data of GPT-4, which might lead to an overestimation of the model's capabilities. In the context of our work, our prompt might just be reminding LLMs of a task they have already seen, whereas on an unseen task, the impact of demonstrations might be different. We do not examine the impact of potential data contamination in our paper and leave an exploration of this to future work.

\paragraph{Other demonstration choices.}
In this work, we choose to use demonstrations that are in the same languages as the test sample, due to our focus on evaluating inherent multilingual abilities of LLMs, as explained in Section~\ref{subsec:multilingual_prompting}.
However, using English demonstrations for cross-lingual transfer or translating test samples into English has its own practical value for NLP applications.
Additionally, it is worth exploring selecting demonstrations from a mixture of languages.
Expanding our study to more setups would provide additional insights into multilingual and cross-lingual LLM abilities.

\paragraph{Other prompting methods.} In Section~\ref{sec:prompt}, we only experiment with manually augmented templates to illustrate how the choice of template can reduce the effectiveness of demonstrations. There is a broad literature on prompt engineering and prompt sensitivity~\citep{white2023prompt, gonen-etal-2023-demystifying}, suggesting that it is plausible that another prompt could reduce the gap between few-shot and zero-shot performance even further.
Chain-of-thought (CoT) prompting is another approach with promising multilingual abilities~\citep{shi2023language, huang-etal-2023-languages} that might affect our findings.
Our manually-augmented templates are intended only as a starting point for further analysis, which we leave to future work.

\paragraph{Beyond automatic evaluation.} When examining model responses, we noticed some cases where a correct answer as evaluated by a human was not fully captured by automatic evaluation metrics. Human evaluation is time-consuming, expensive, and hard to source for the wide range of languages that we explore in our work. Another option is LLM evaluation, which is becoming increasingly popular~\citep{fu2023gptscore, chan2023chateval}, but is also an expensive approach. More importantly, we have no guarantees about LLMs' multilingual capabilities. As a trade-off between cost and evaluation quality, we stick to automatic evaluation in our work for all tasks and languages.

\section*{Acknowledgments}
We thank Matan Eyal for his valuable feedback. Our use of Microsoft Azure is sponsored by the Microsoft Accelerating Foundation Models Research (AFMR) program. Miaoran Zhang and Marius Mosbach received funding from the DFG (German Research Foundation) under project 232722074, SFB 1102. Vagrant Gautam and Jesujoba O. Alabi were supported by the BMBF's (German Federal Ministry of Education and Research) SLIK project under the grant 01IS22015C.

\bibliography{anthology,custom}

\newpage
\appendix
\onecolumn
\section{Experimental details}
\label{sec:appendix_setup}

\subsection{Tasks and datasets}
\label{sec:appendix_tasks_datasets}
We conduct experiments on 9 multilingual datasets with a wide coverage of tasks and languages, as shown in Table~\ref{tab:benchmark}. All datasets are public research datasets and our experiments are consistent with their intended use, i.e., NLP evaluation. For the machine translation dataset MAFAND, English serves as the pivot language and there are two translation directions: en-xx (i.e., translating from English to another language) and xx-en (i.e., translating from another language to English). As the black-box training data of OpenAI APIs that we used is up to September 2021, we include the dataset release date in the table which can be taken as a clue to the severity of dataset contamination. 

\definecolor{Color}{gray}{0.9}

\begingroup
\renewcommand{\arraystretch}{1.2} %

\begin{table*}[htb]
\resizebox{\linewidth}{!}{
    \begin{tabular}{lllcc}
    \toprule
    \textbf{Dataset}      & \textbf{Task}          &\textbf{Languages}    & \textbf{|Lang.|}& \textbf{Release Date}  \\ 
    \midrule
    XNLI      & natural language inference  & English, German, Russian, French, Spanish, Chinese, Vietnamese,  &15 & 2019.09 \\
    &&Turkish, Arabic, Greek, Thai, Bulgarian, Hindi, Urdu, Swahili \\
    \rowcolor{Color}IndicXNLI  & natural language inference  & Hindi, Bengali, Tamil, Marathi, Malayalam, Telugu, Kannada, Punjabi,
    & 11& 2022.04\\
    \rowcolor{Color}&&  Oriya, Assamese, Gujarati && \\
    PAWS-X     & paraphrase identification   & English, German, Japanese, French, Spanish, Chinese, Korean &7 & 2019.08 \\
    \rowcolor{Color}XCOPA      & commonsense reasoning   &  Chinese, Italian,    Vietnamese, Indonesian, Turkish, Thai, Estonian,& 11& 2020.04\\
    \rowcolor{Color}&& Tamil, Swahili, Haitian, Quechua  && \\
    XStoryCloze  & commonsense reasoning   &English, Russian, Spanish, Chinese, Indonesian, Arabic, Hindi, & 11 & 2023.05 \\
    &&  Basque, Telugu, Burmese, Swahili \\
    \rowcolor{Color}AfriSenti    & sentiment analysis & Swahili, Amharic, Hausa, Kinyarwanda, Yoruba, Tigrinya, Igbo, Oromo,  & 14& 2023.05 \\
    \rowcolor{Color}&& Moroccan Arabic, Algerian Arabic, Nigerian Pidgin, Mozambican Portuguese, &&\\
    \rowcolor{Color}&&Tsonga, Twi &&\\
    XQuAD        & extractive QA    &English, German, Russian, Spanish, Chinese, Vietnamese, Turkish, Greek, & 12  & 2019.10 \\
    && Romanian, Thai, Hindi \\
    \rowcolor{Color}TyDiQA-GoldP & extractive QA     &English, Russian, Indonesian, Korean, Arabic, Finnish, Bengali, Telugu, Swahili& 9 & 2020.02\\
    MAFAND & machine translation & Amharic, Hausa, Kinyarwanda, Luganda, Luo, Chichewa, Nigerian Pidgin, & 14& 2022.06  \\
    &&Shona, Swahili, Setswana, Twi, Xhosa, Yoruba, Zulu \\
    \bottomrule
    \end{tabular}}
\caption{Multilingual benchmarking datasets.}
\label{tab:benchmark}
\end{table*}

\endgroup

\vspace{-3mm}

\subsection{In-context learning}
\label{sec:appendix_prompting_setup}
We sample few-shot demonstrations from the validation set and evaluate the test set. For datasets without a test data split (XStoryCloze and TyDiQA), we sample few-shots from the train set and evaluate the validation set. Since XQuAD only has a validation data split, we utilize it for both demonstration sampling and evaluation, ensuring that the test sample itself is not included in its demonstrations. For chat models (Llama 2-Chat, GPT-3.5, and GPT-4), we limit the test sample size to a maximum of $200$ in order to reduce inference expenses and ensure a fair comparison.   

We use GPT-3 style prompting templates for XGLM and Llama 2 as shown in Table~\ref{tab:template_xglm}. The templates for BLOOMZ and mT0 are shown in Table~\ref{tab:template_bloomz}. For Llama 2-Chat, GPT-3.5 and GPT-4, default templates are shown in Table~\ref{tab:template_openai} and task instructions are used to assign a system role to the model. Inspired by \citet{lai-etal-2023-chatgpt} and \citet{li2023large}, where emotional stimuli are able to enhance LLM understanding, we design formatting-focused templates (discussed in Section \ref{sec:prompt}) to reinforce LLM to generate formatted outputs that are easier to evaluate automatically, as shown in Table~\ref{tab:template_openai_1}.

\subsection{Implementation}
\label{sec:appendix_implementation}
Our codebase is adapted from OpenICL~\citep{wu-etal-2023-openicl}. We use int8bit model quantization\footnote{In our preliminary experiments, we found that int8 quantization led to a performance degradation of 1-2\% on a few classification datasets with Llama 2 and XGLM. Since this degradation is consistent across different setups, we believe that it would not affect our overall findings.} for all models except OpenAI models. Experiments are conducted using a single NVIDIA A100-80GB GPU. As models have a maximum context length, we preserve complete demonstrations that can fit within the context window. We employ greedy decoding for model generation. For chat models, the maximum new token is set to $50$, while for machine translation, it is set to $100$. For other models, the maximum new token is set to $20$, while for machine translation, it is set to $50$. We use three seeds ($0, 33, 42$) in our experiments, and the single-seed results for BLOOMZ and mT0 are obtained with the seed $0$.

\section{More results for varying numbers of demonstrations}
\label{sec:appendix_n_shot}
In this section, we provide supplemental results for Section~\ref{sec:n_demo}.

\subsection{Results for BLOOMZ and mT0}
\label{sec:appendix_bloomz_mt0}

\begin{figure*}[h]
    \centering
    \includegraphics[width=\linewidth]{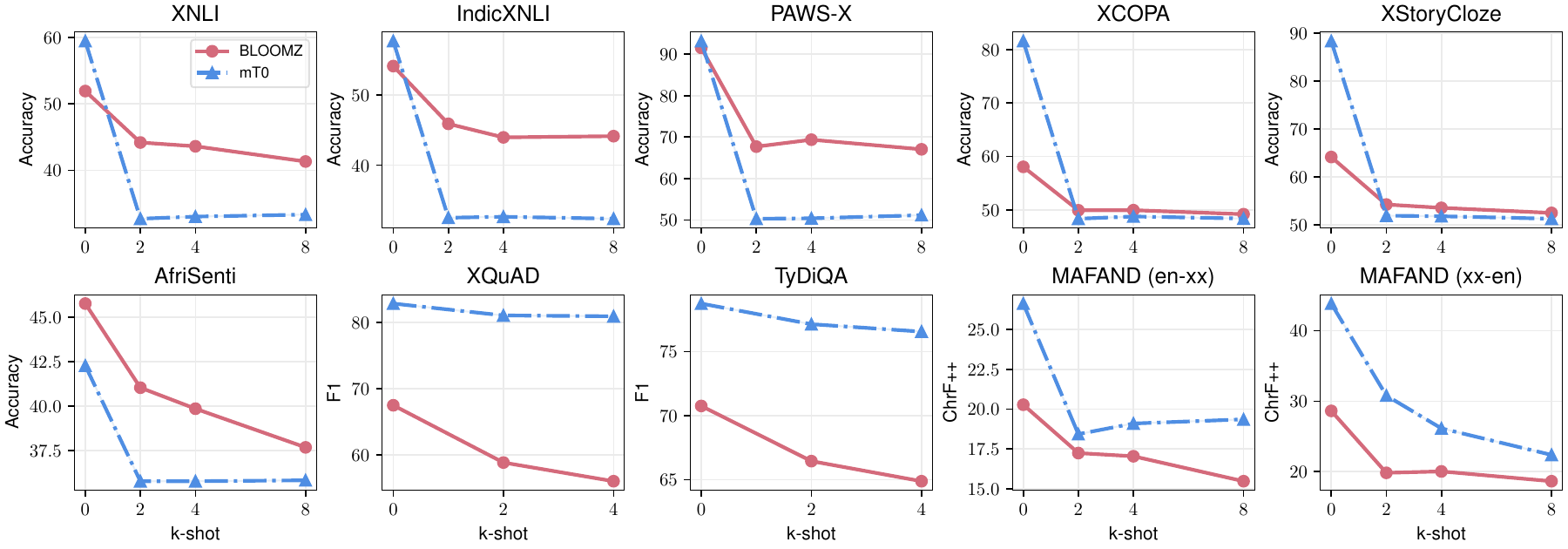}
    \caption{Average performance across languages for BLOOMZ and mT0 with different numbers of demonstrations. The results are obtained with a single random seed. Note that PAWS-X, XQuAD and TyDiQA are included in the instruction-tuning datasets of BLOOMZ and mT0.}
    \label{fig:n_shot_bloomz}
\end{figure*}

In addition to the $5$ models (base models and chat models) we discussed in the main content, we also experiment with two instruction-tuned models: BLOOMZ and mT0~\citep{muennighoff-etal-2023-crosslingual}. The results for varying numbers of random demonstrations are shown in Figure~\ref{fig:n_shot_bloomz}. In line with findings from~\citet{asai2023buffet}, we observe significant performance degradation when using demonstrations compared to zero-shot learning in all cases. This decline can be attributed to their training scheme, where models are fine-tuned on a large collection of existing datasets in a zero-shot manner. In contrast, several studies~\citep{chen-etal-2022-meta, wang-etal-2022-super} focus on enhancing the in-context learning ability of LLMs by incorporating demonstrations into their training process. This suggests that we should be careful in model selection for in-context learning research and take the model training process into consideration.
 
\subsection{Results for individual languages}
\label{sec:appendix_n_shot_languages}
The language-specific results for each task are shown in Figure~\ref{fig:n_shot_all}. The order of languages follows their data ratio in the CommonCrawl corpus\footnote{\url{http://commoncrawl.org}} from high-resource to low-resource. We observe large variations in model performance across different languages. For instance, there exists a large performance disparity between English and Urdu in XNLI. In XCOPA, the performance of Quechua is significantly worse compared to other languages.

\begin{figure*}[h]
    \begin{subfigure}[]{\linewidth}
    \centering
    \includegraphics[width=0.95\linewidth]{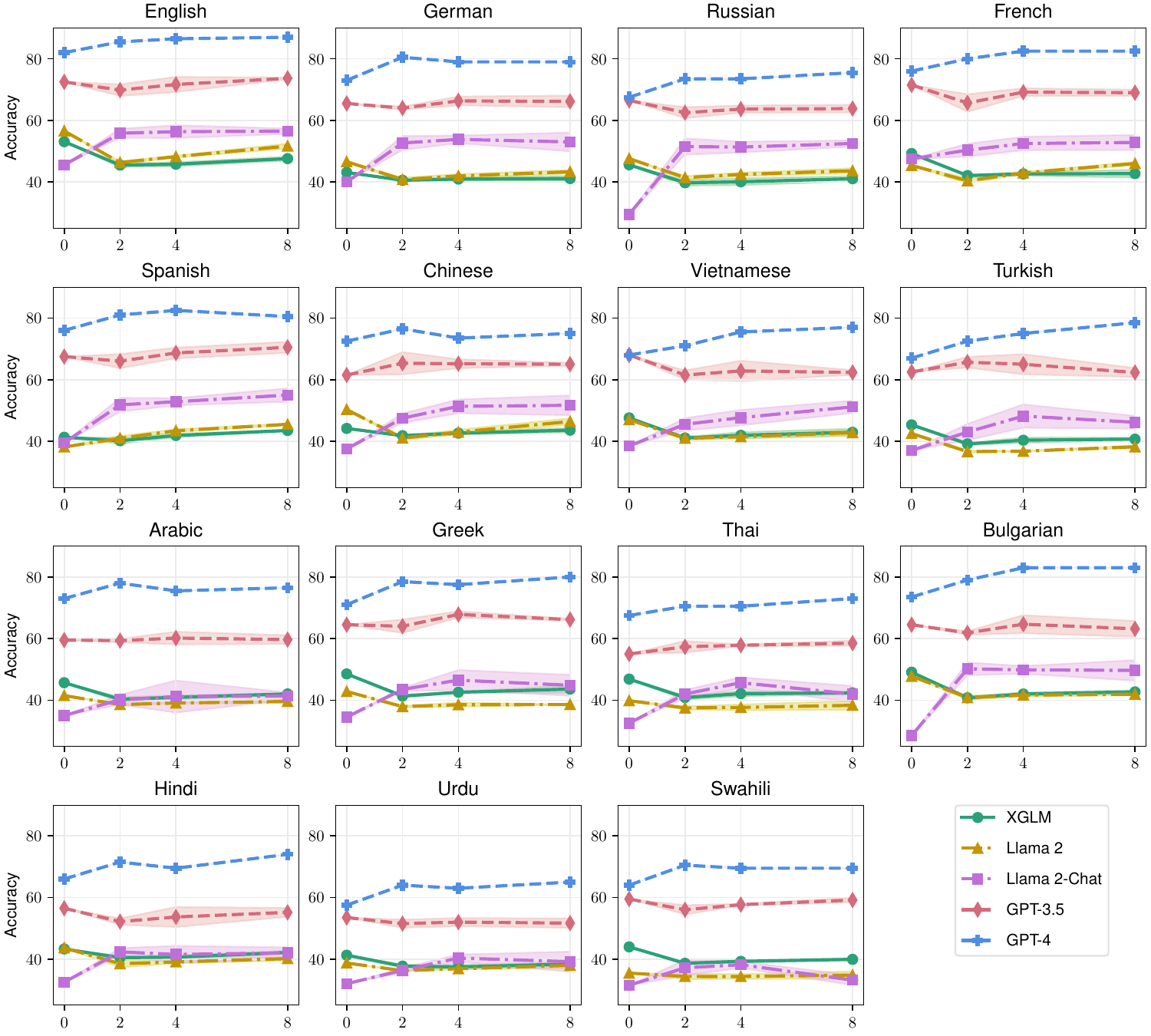}
    \caption{XNLI}
    \label{fig:n_shot_xnli}
    \end{subfigure}

    \begin{subfigure}[]{\linewidth}
    \centering
    \includegraphics[width=0.95\linewidth]{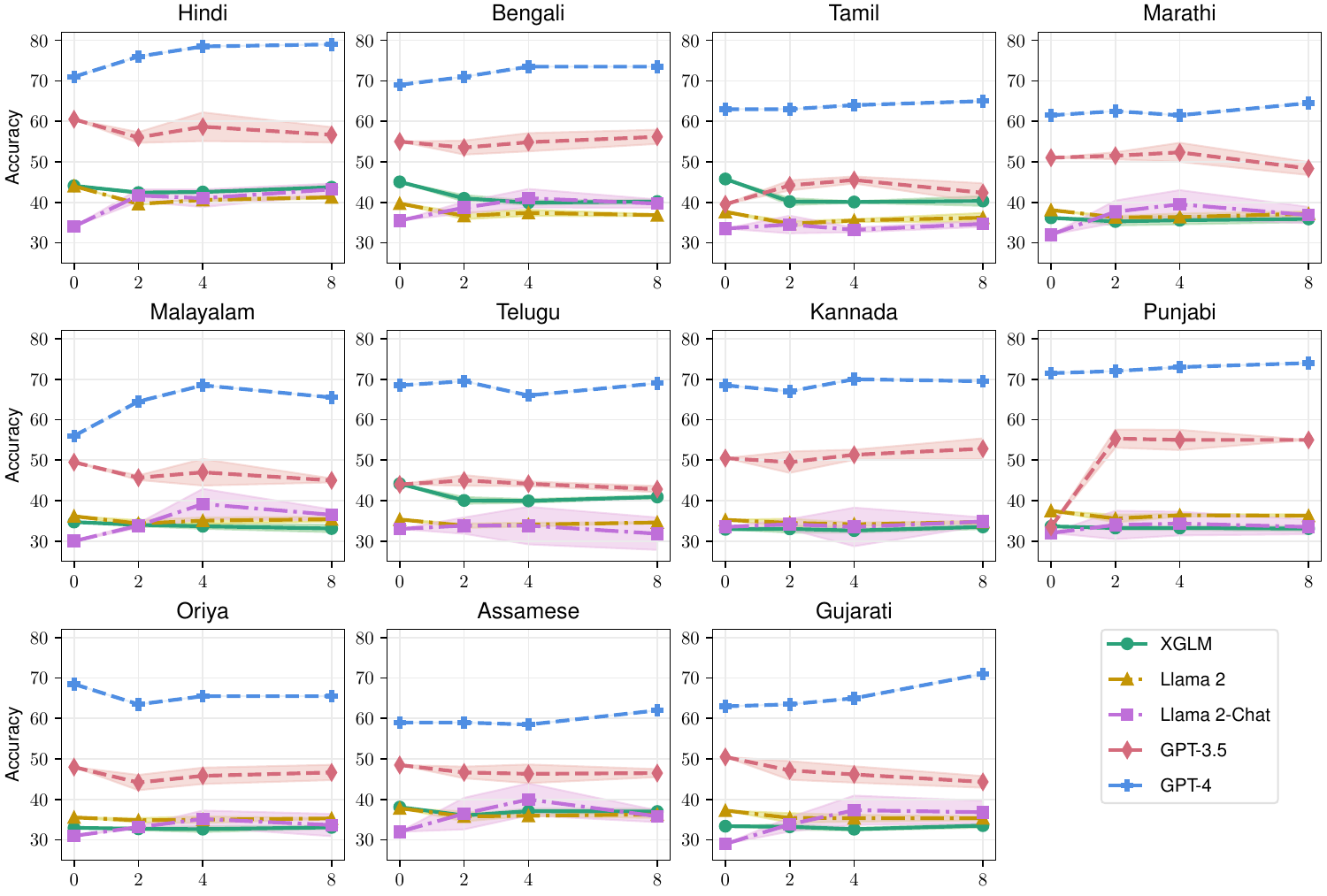}
    \caption{IndicXNLI}
    \label{fig:n_shot_indicxnli}
    \end{subfigure}
\end{figure*}

\begin{figure*}[h]
    \ContinuedFloat
    
    \begin{subfigure}[]{\linewidth}
    \centering
    \includegraphics[width=0.95\linewidth]{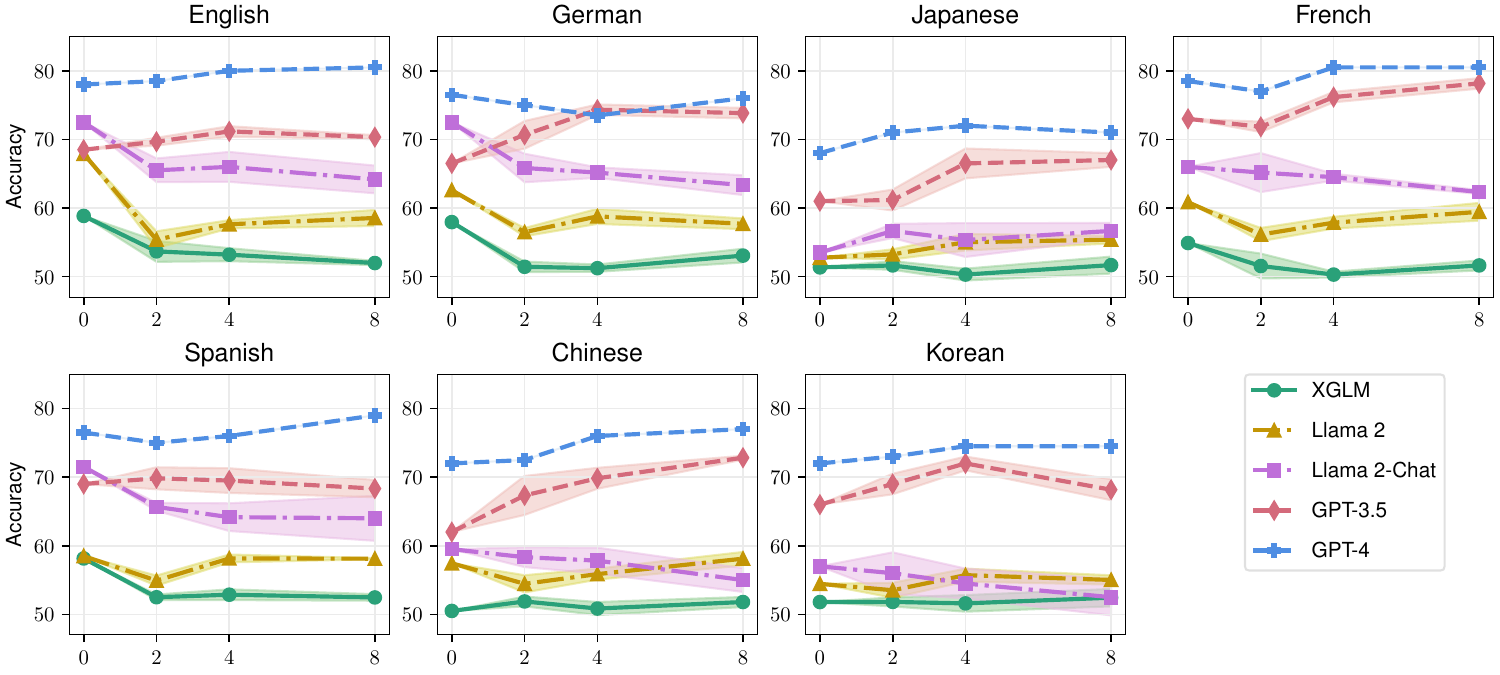}
    \caption{PAWS-X}
    \label{fig:n_shot_pawsx}
    \end{subfigure}

    \vspace{5em}

    \begin{subfigure}[]{\linewidth}
    \centering
    \includegraphics[width=0.95\linewidth]{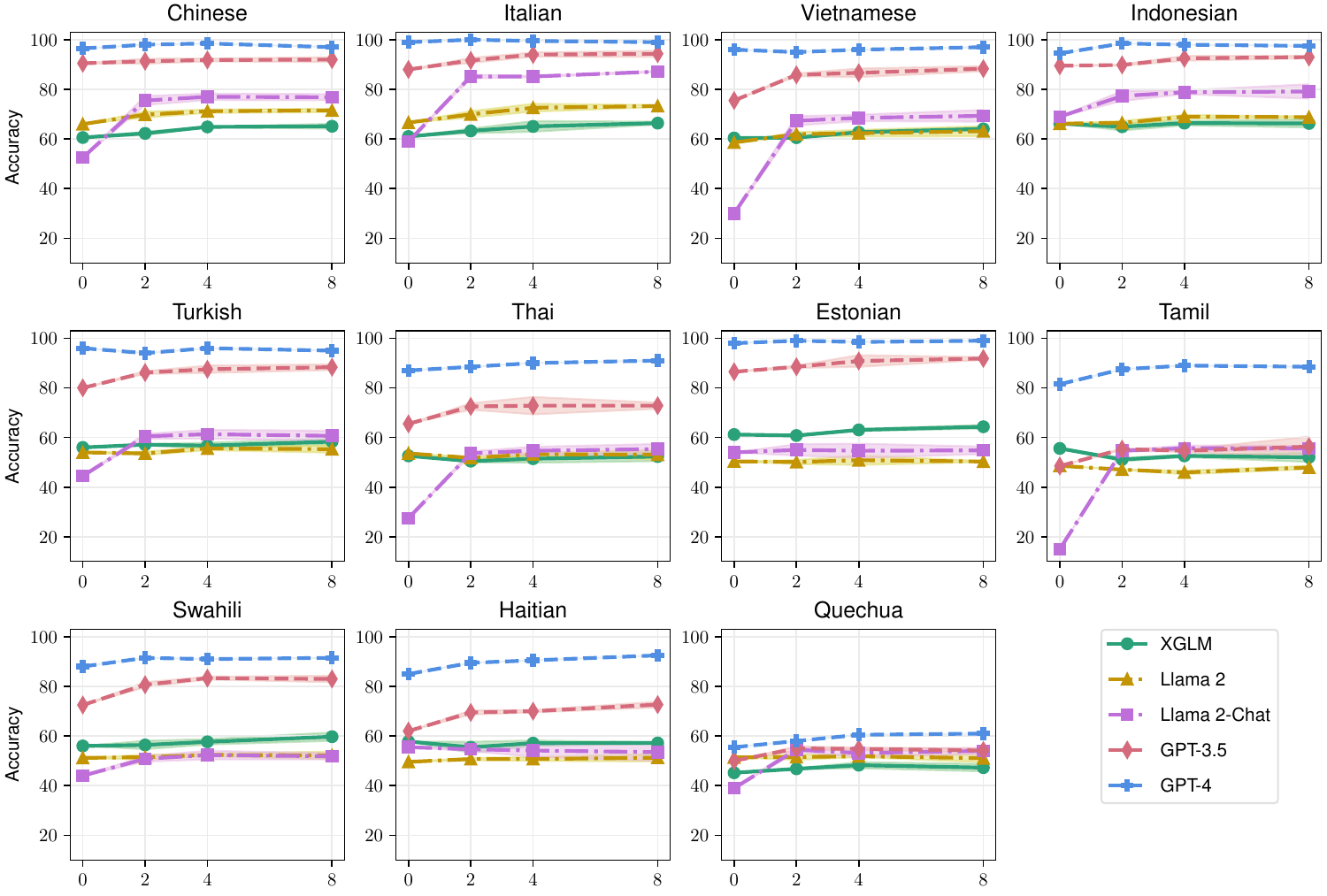}
    \caption{XCOPA}
    \label{fig:n_shot_xcopa}
    \end{subfigure}
    
\end{figure*}

\begin{figure*}[h]
    \ContinuedFloat
    \begin{subfigure}[]{\linewidth}
    \centering
    \includegraphics[width=0.95\linewidth]{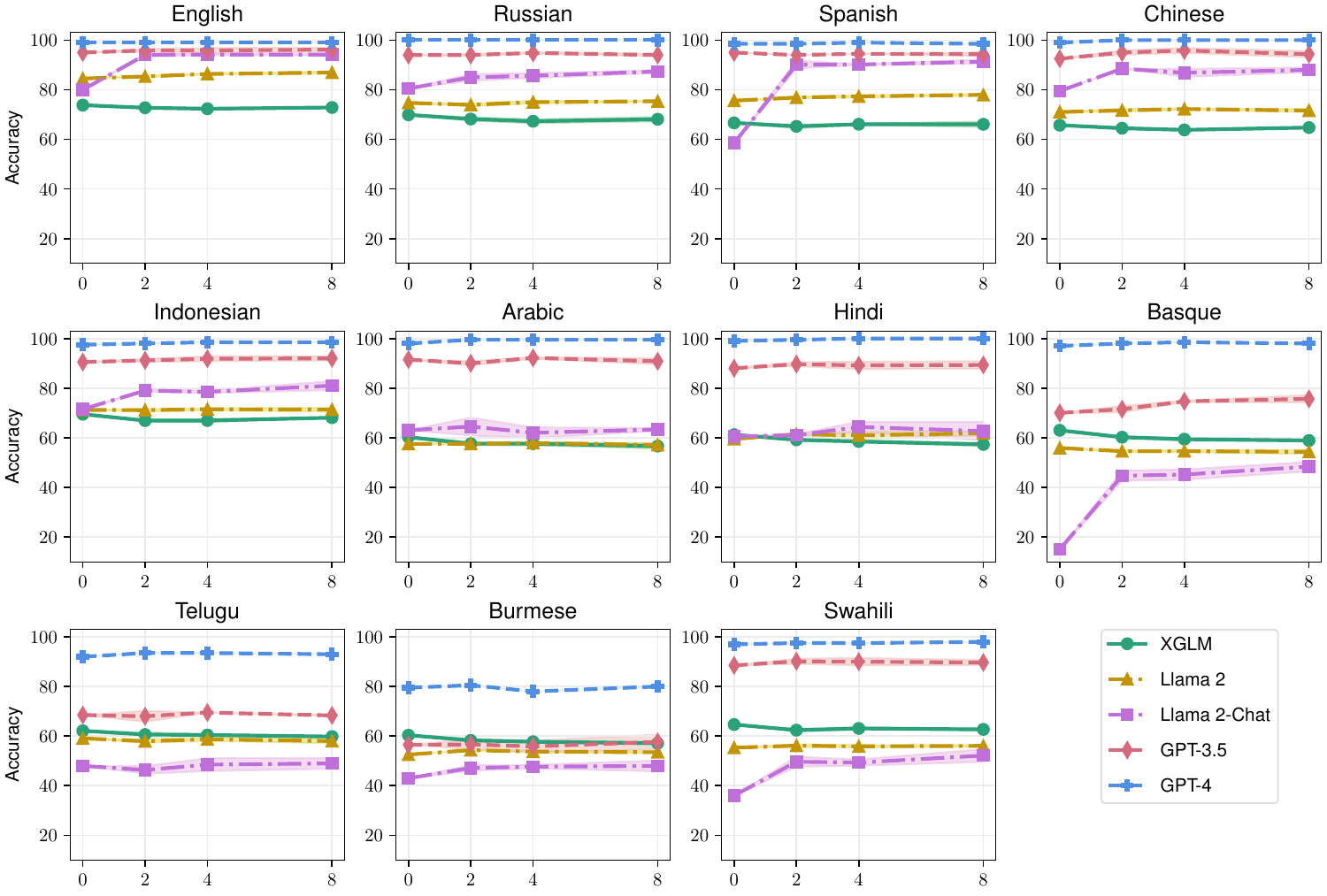}
    \caption{XStoryCloze}
    \label{fig:n_shot_xstorycloze}
        
    \end{subfigure}

    \begin{subfigure}[]{\linewidth}
    \centering
    \includegraphics[width=0.95\linewidth]{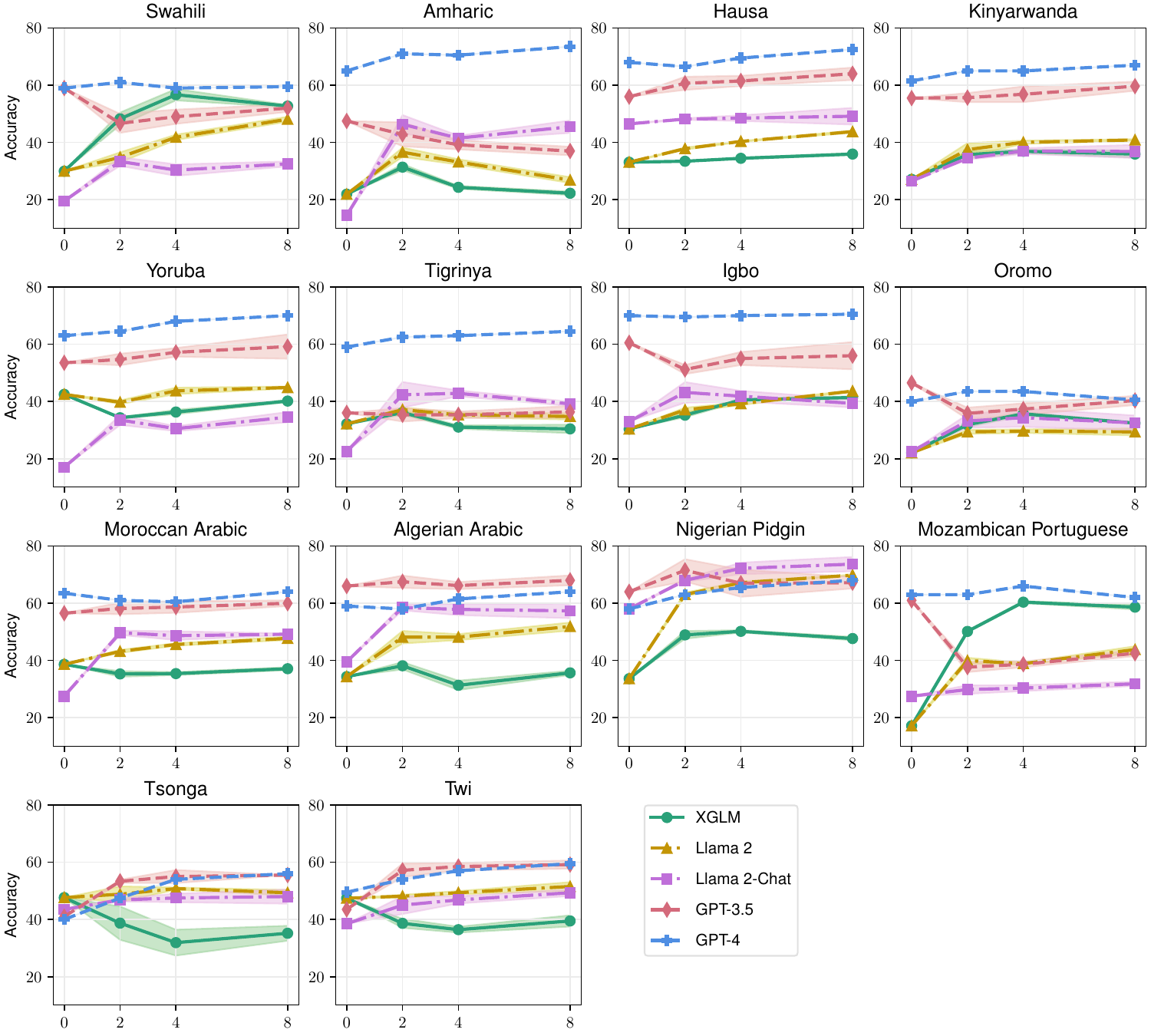}
    \caption{AfriSenti}
    \label{fig:n_shot_afrisenti}
    \end{subfigure}
    
\end{figure*}

\begin{figure*}[h]
    \ContinuedFloat
    
    \begin{subfigure}[]{\linewidth}
    \centering
    \includegraphics[width=0.95\linewidth]{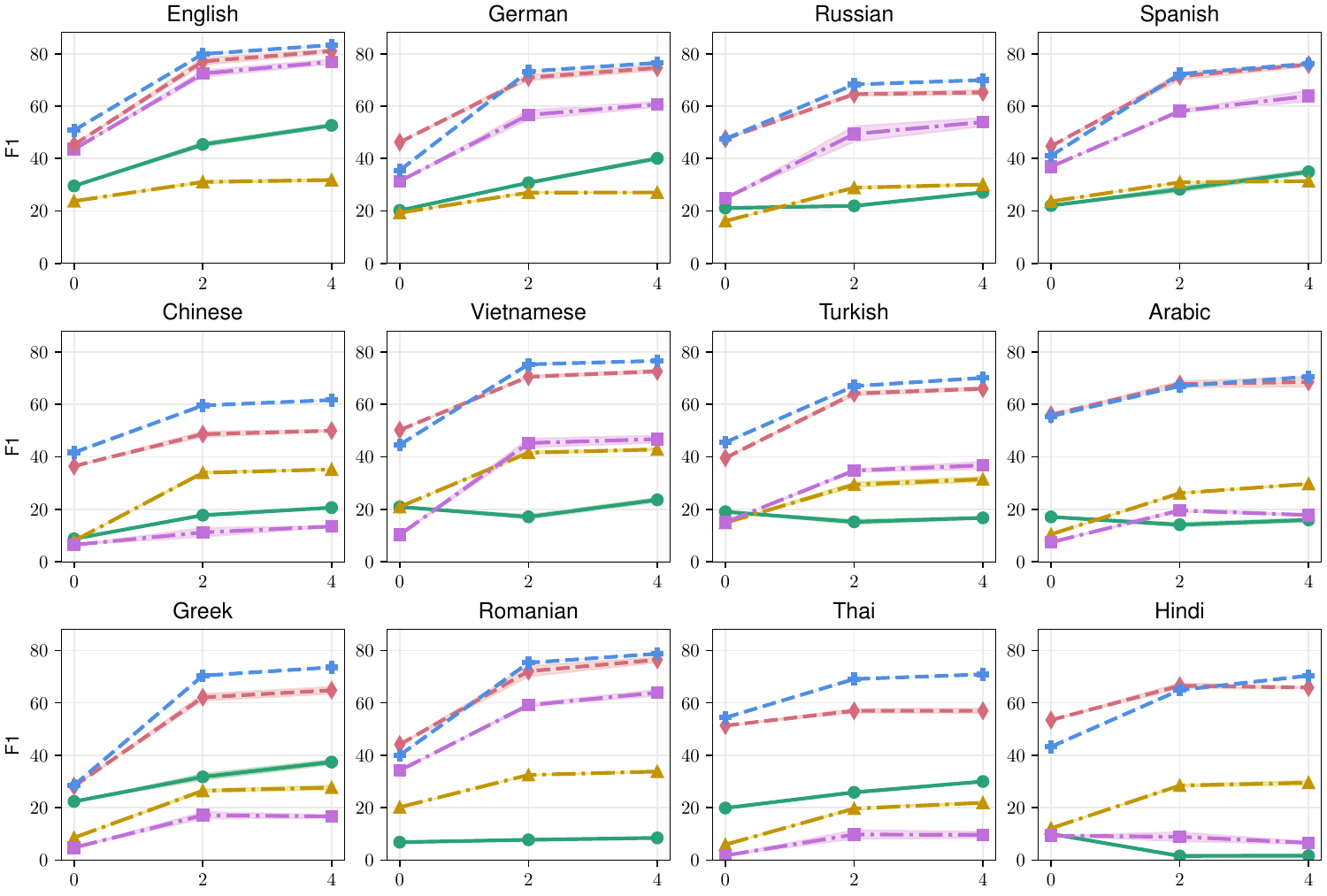}
    \caption{XQuAD}
    \label{fig:n_shot_xquad}
    \end{subfigure}

    \begin{subfigure}[]{\linewidth}
    \centering
    \includegraphics[width=0.95\linewidth]{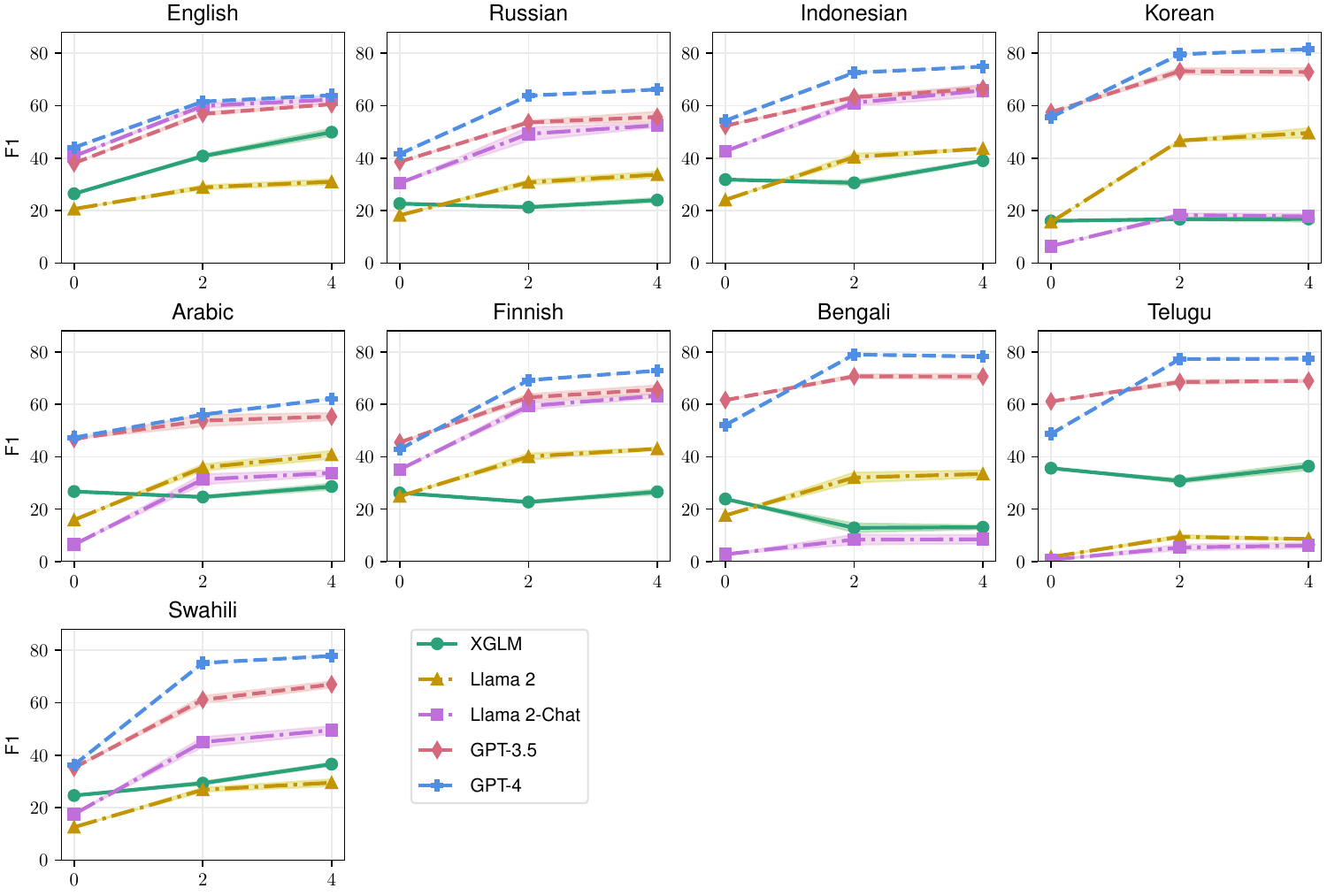}
    \caption{TyDiQA}
    \label{fig:n_shot_tydiqa}
    \end{subfigure}
\end{figure*}

\begin{figure*}
    \ContinuedFloat
    \begin{subfigure}[]{\linewidth}
    \centering
    \includegraphics[width=0.95\linewidth]{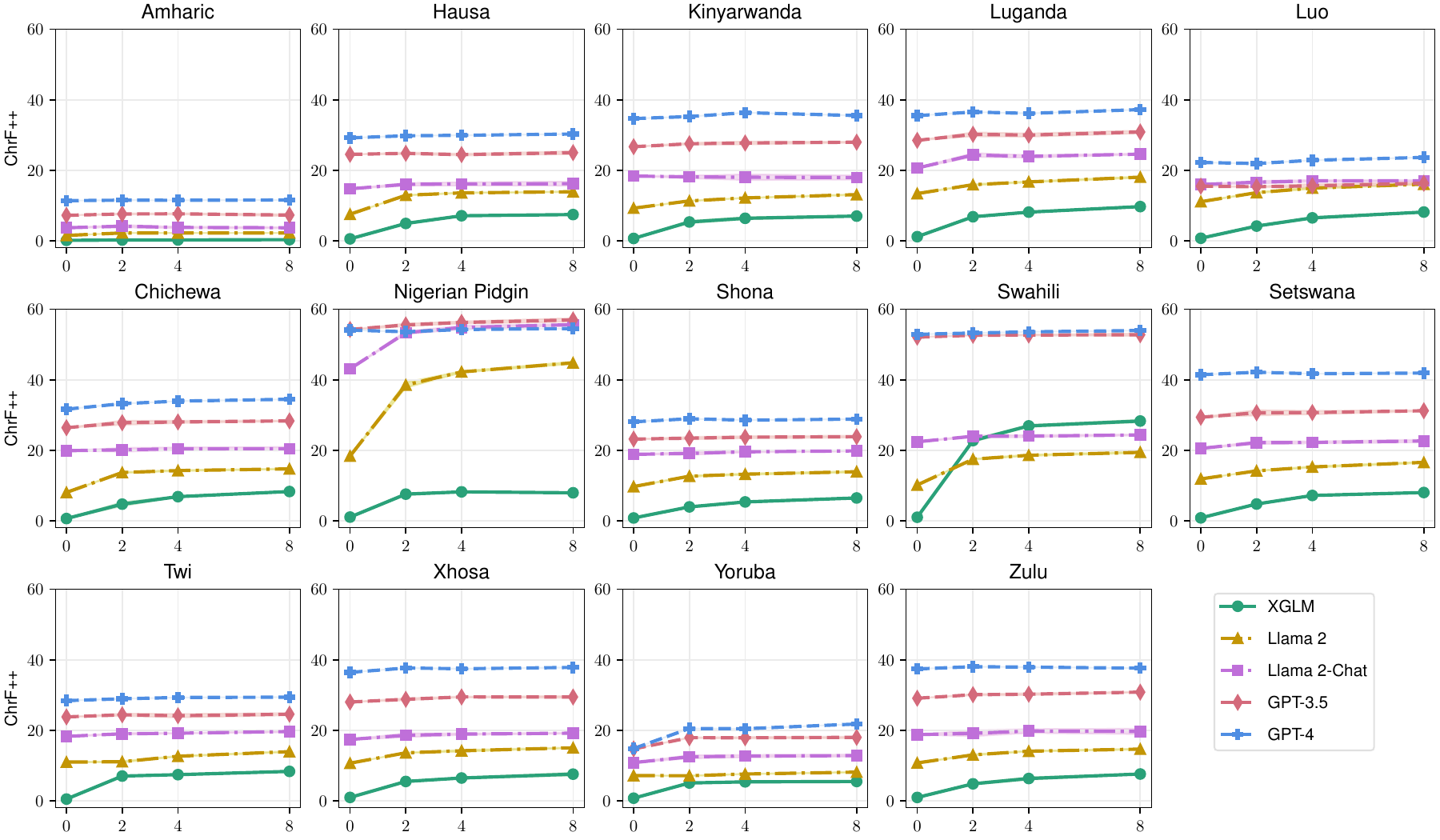}
    \caption{MAFAND (en-xx)}
    \label{fig:n_shot_mafand_e2t}
    \end{subfigure}

    \vspace{1em}
    
    \begin{subfigure}[]{\linewidth}
    \centering
    \includegraphics[width=0.95\linewidth]{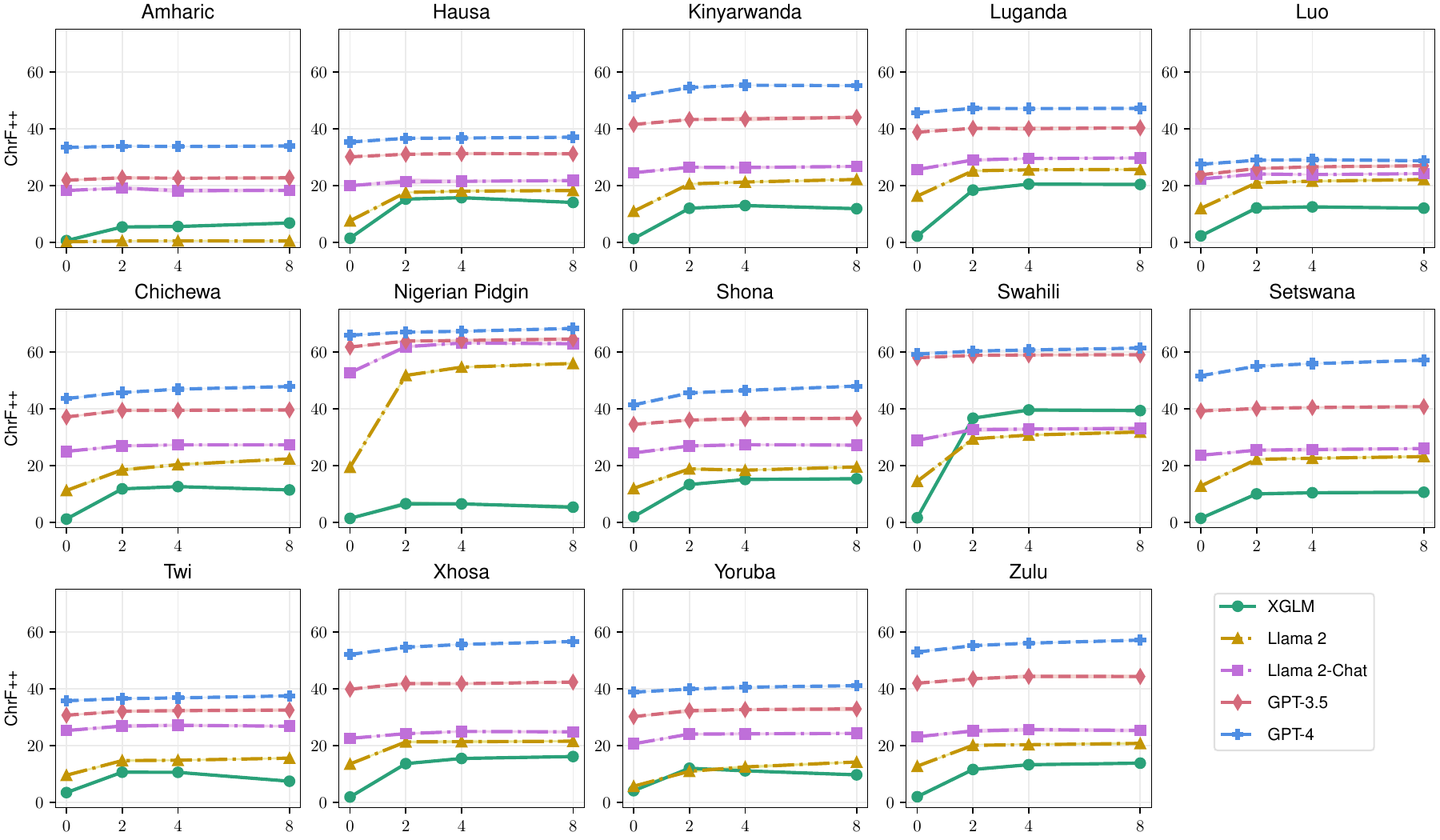}
    \caption{MAFAND (xx-en)}
    \label{fig:n_shot_mafand_t2e}
    \end{subfigure}
    \caption{Language-specific performance for both classification and generation tasks with different numbers of demonstrations. We average and report standard deviations over 3 seeds for all models except GPT-4.}
    \label{fig:n_shot_all}
\end{figure*}

\section{More results for ablating the quality of demonstrations}
\label{sec:appendix_quality}
In this section, we provide supplemental results for Section~\ref{sec:quality}.

\subsection{Performance of different types of demonstrations}
\label{sec:appendix_demo_performance}
In Table~\ref{tab:demos}, we show the model performance of three types of demonstrations, as well as the zero-shot performance for comparative analysis. As we notice, top-k selection may not always be the optimal choice, given the considerable effort in optimizing demonstrations. For QA, XGLM and Llama 2's abilities in solving this task almost collapse with corrupted labels. However, for chat models, demonstrations with corrupted labels can achieve comparable performance with ground truth labels and largely improve the zero-shot performance. Overall, the base models are more sensitive to the type of demonstrations than chat models. 

\subsection{Results for individual languages}
\label{sec:appendix_quality_languages}
In Figure~\ref{fig:all_demo}, we show the language-specific results for each task, in which we can see language discrepancies with different types of demonstrations.

\begin{figure}[h]

    \centering
    \begin{subfigure}[]{0.49\textwidth}
    \centering
    \includegraphics[width=\textwidth]{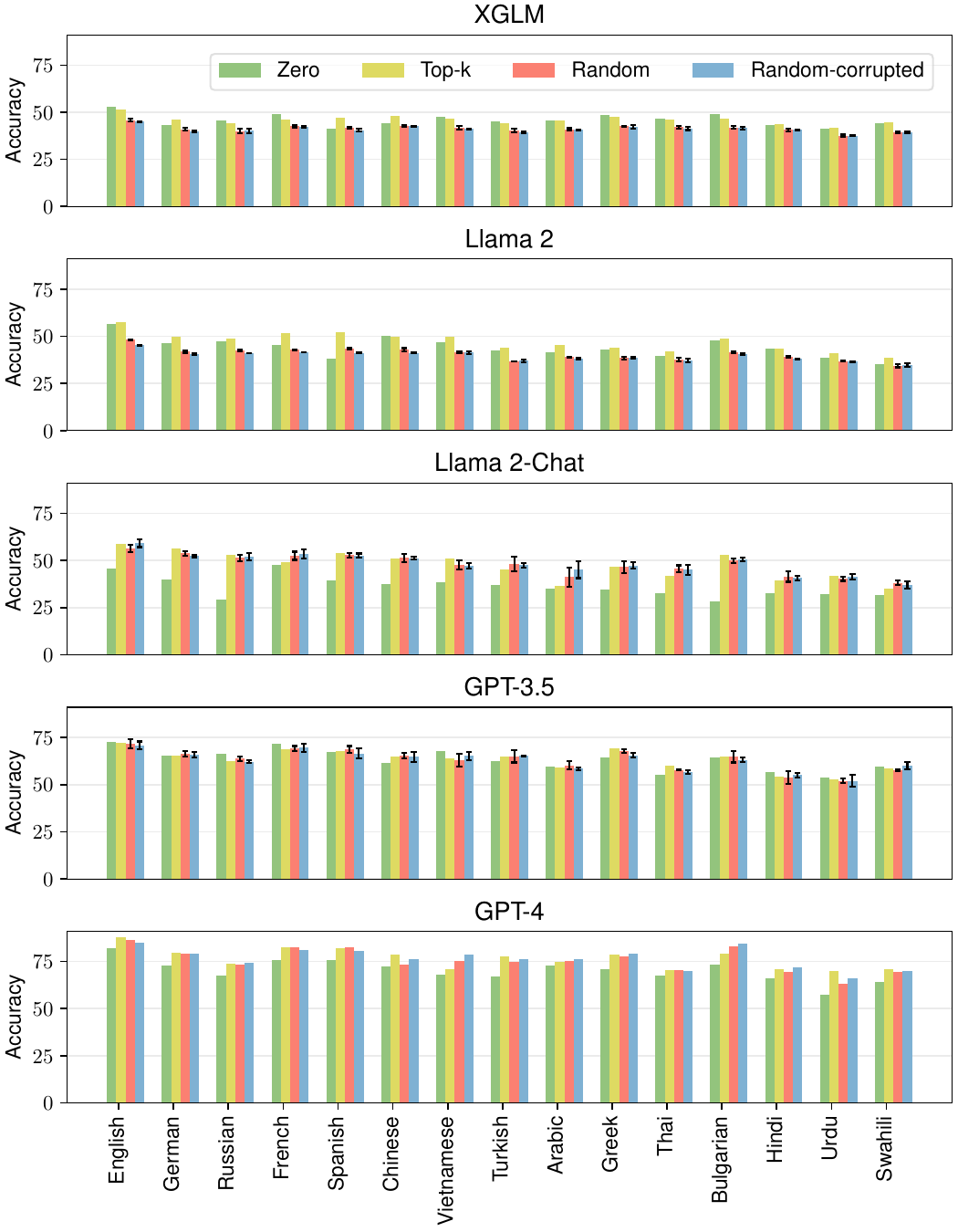}
    \caption{XNLI}
    \label{fig:xnli_demo}
    \end{subfigure}
    \begin{subfigure}[]{0.49\textwidth}
    \centering
    \includegraphics[width=\textwidth]{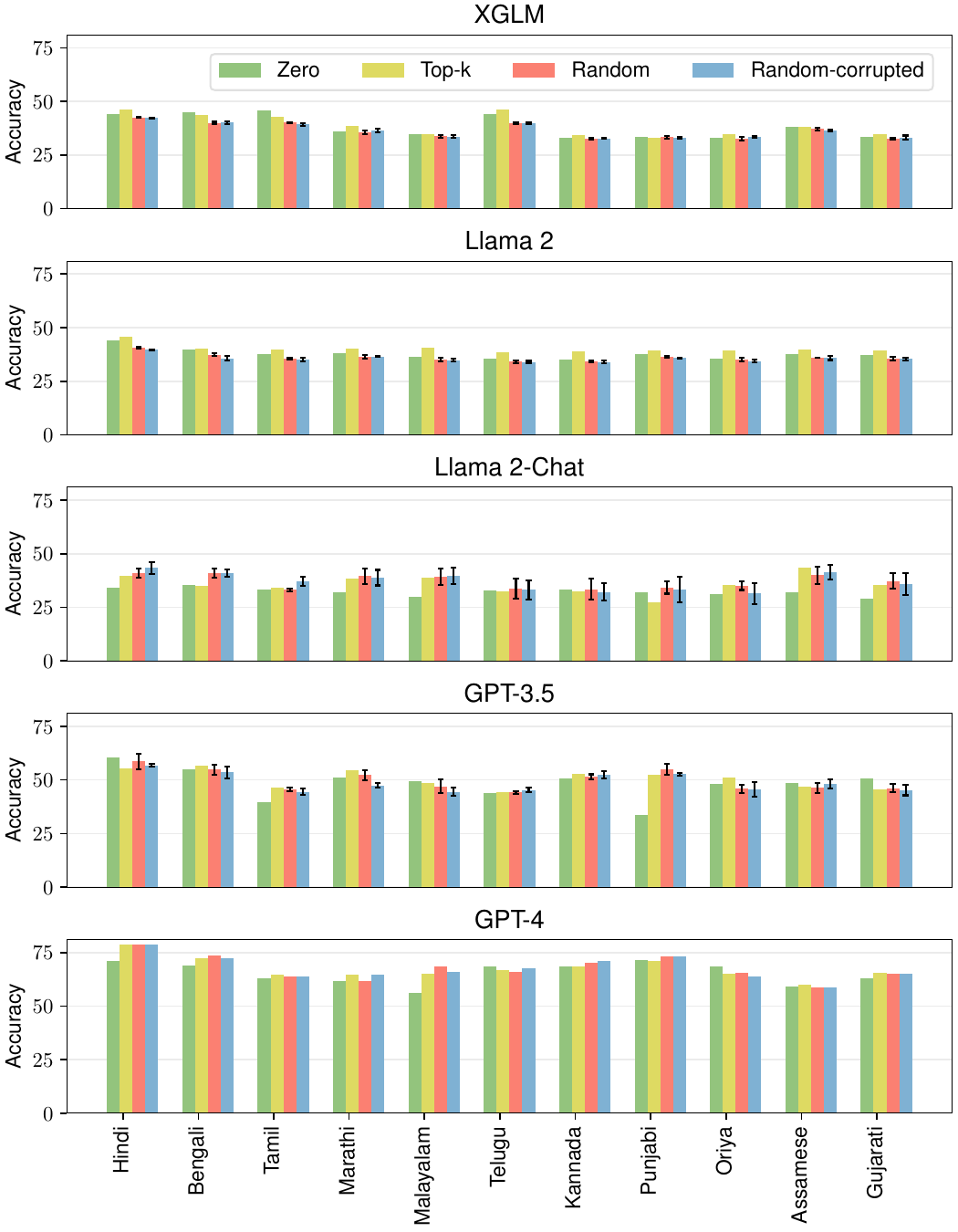}
    \caption{IndicXNLI}
    \label{fig:indicxnli_demo}
    \end{subfigure}

\end{figure}

\begin{figure}[h]
    \centering
    \ContinuedFloat

    \begin{subfigure}[]{0.49\textwidth}
    \centering
    \includegraphics[width=\textwidth]{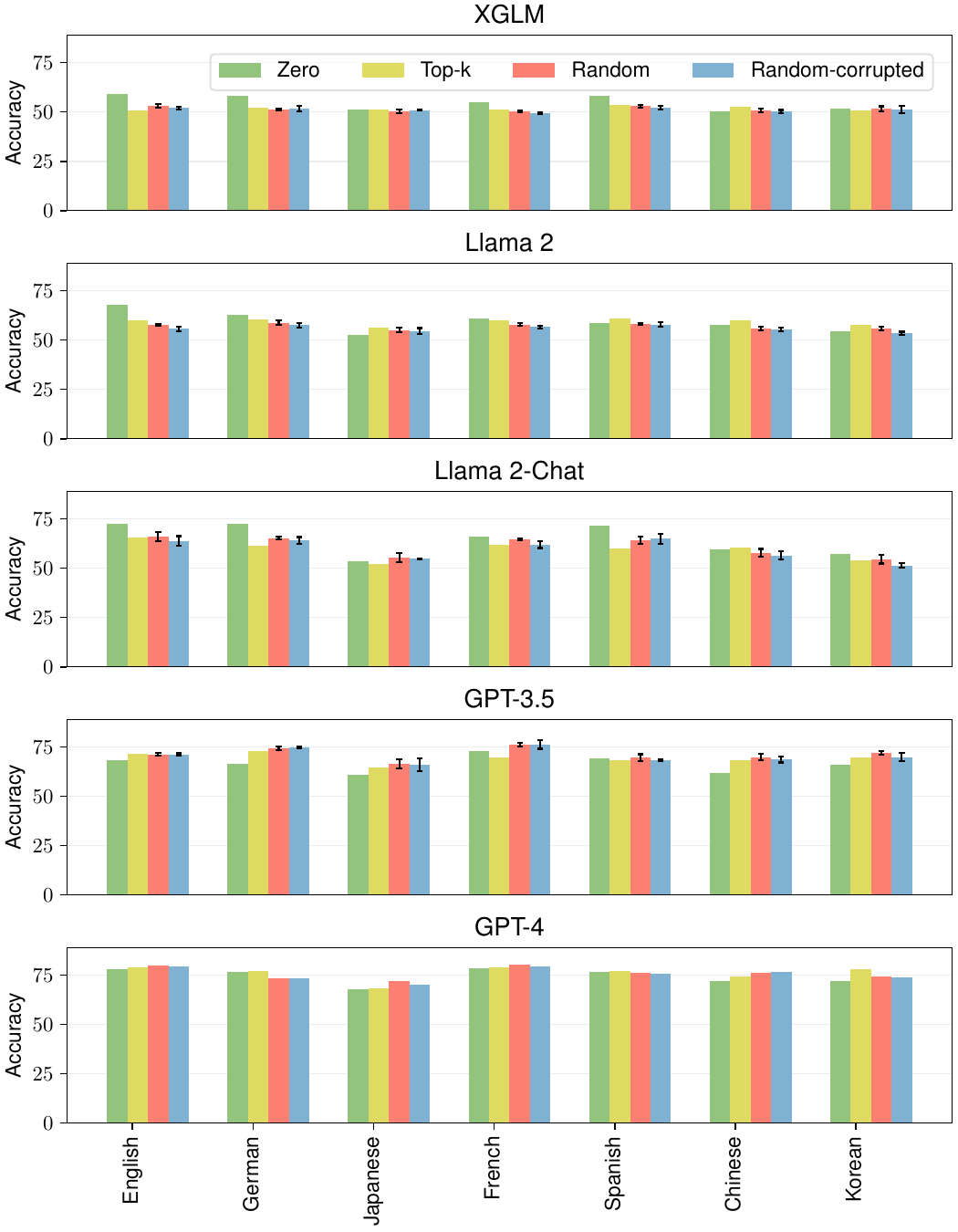}
    \caption{PAWS-X}
    \label{fig:pawsx_demo}
    \end{subfigure}
    \begin{subfigure}[]{0.49\textwidth}
    \centering
    \includegraphics[width=\textwidth]{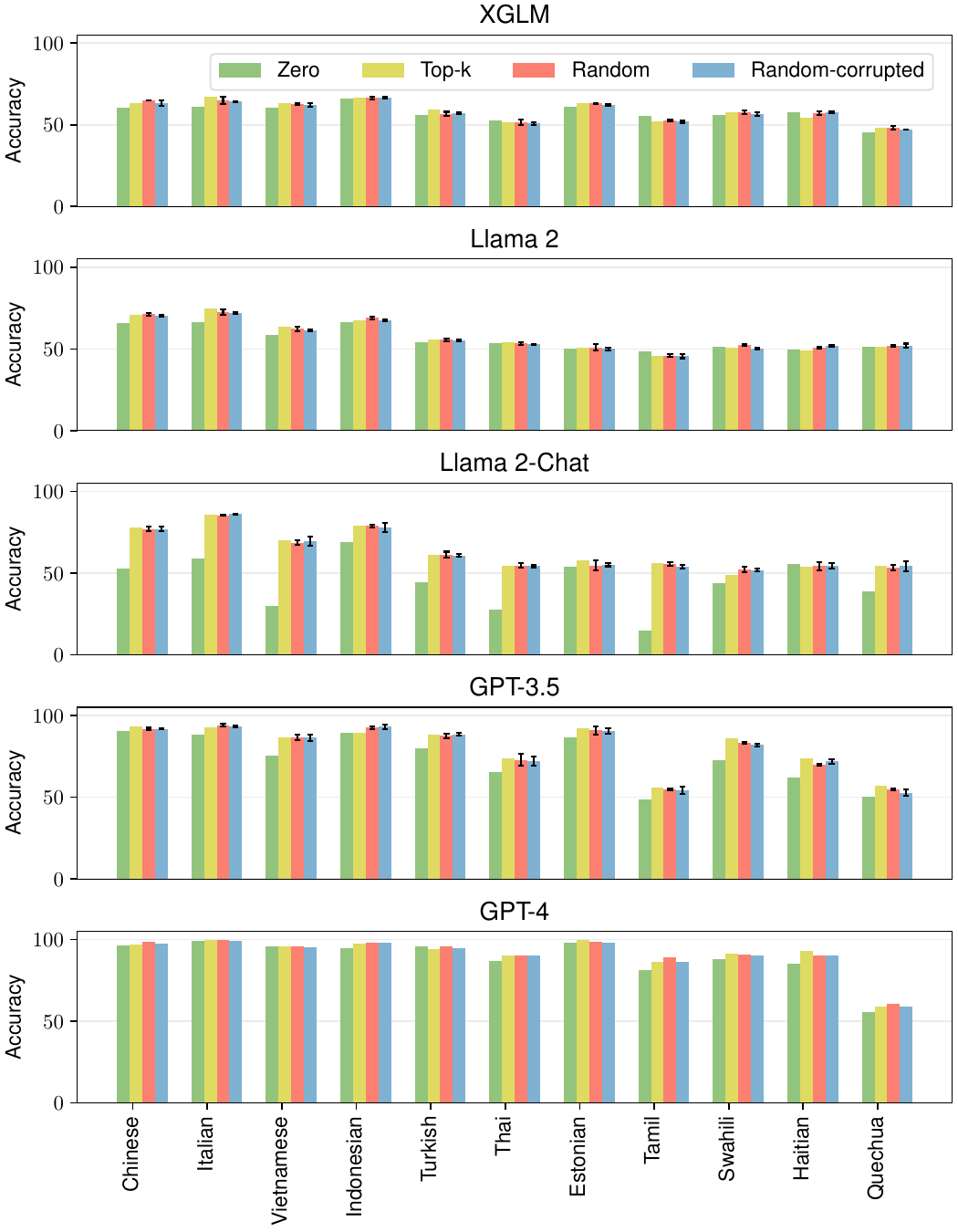}
    \caption{XCOPA}
    \label{fig:xcopa_demo}
    \end{subfigure}
    
\end{figure}

\begin{figure}[h]
    \centering
    \ContinuedFloat

    \begin{subfigure}[]{0.49\textwidth}
    \centering
    \includegraphics[width=\textwidth]{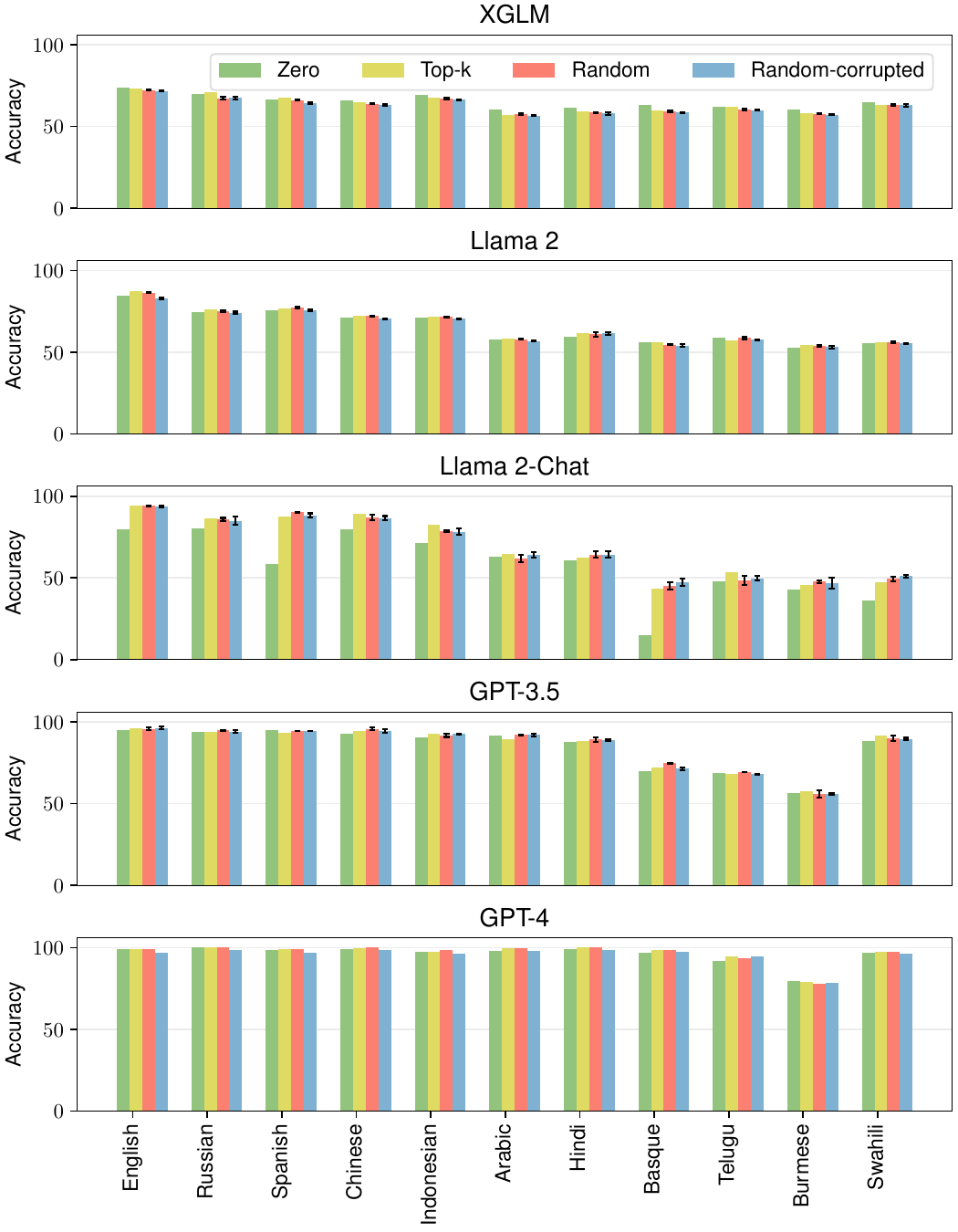}
    \caption{XStoryCloze}
    \label{fig:xstorycloze_demo}
    \end{subfigure}
    \begin{subfigure}[]{0.49\textwidth}
    \centering
    \includegraphics[width=\textwidth]{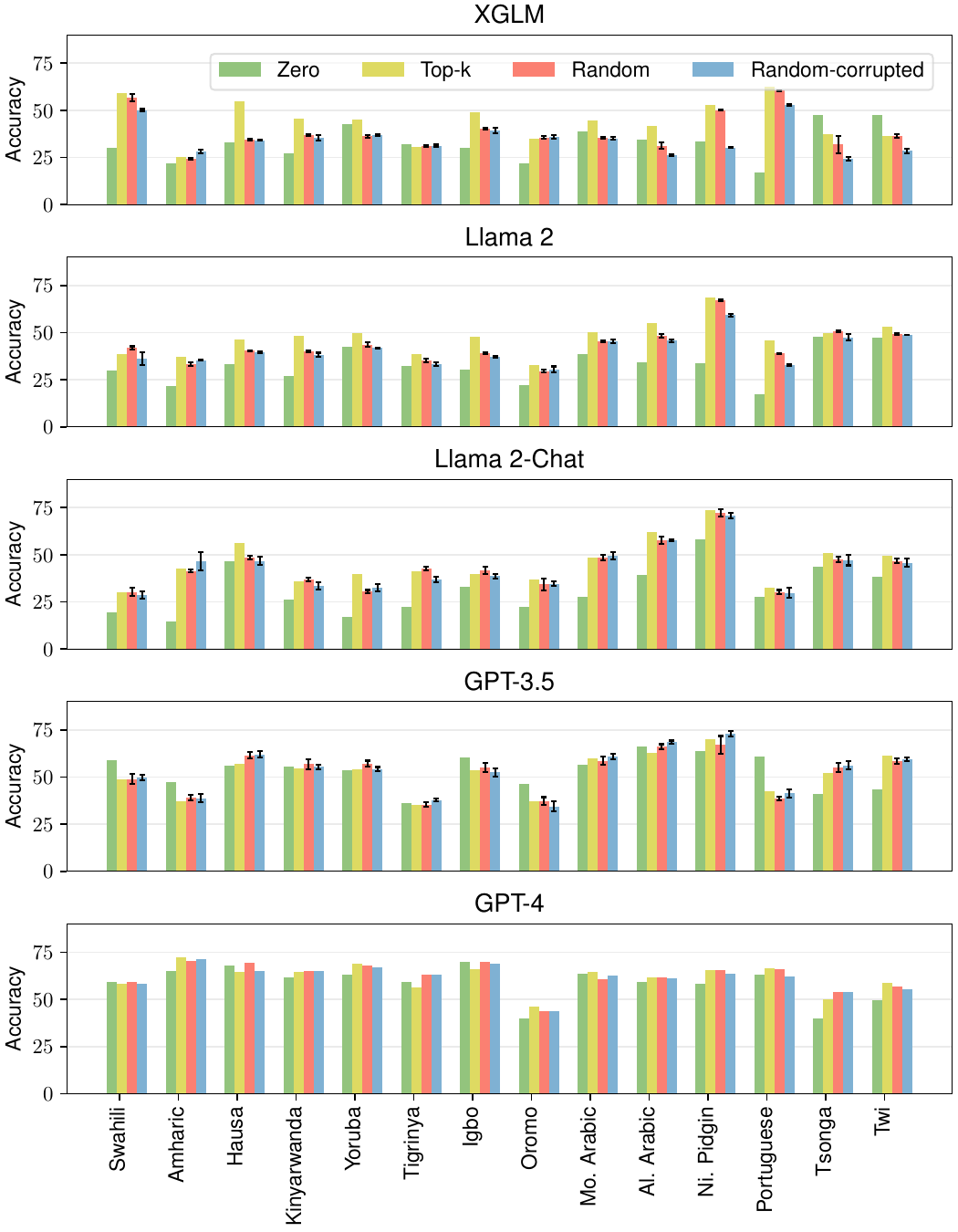}
    \caption{AfriSenti}
    \label{fig:afrisenti_demo}
    \end{subfigure}

\end{figure}

\begin{figure}[h]
    \centering
    \ContinuedFloat

    \begin{subfigure}[]{0.49\textwidth}
    \centering
    \includegraphics[width=\textwidth]{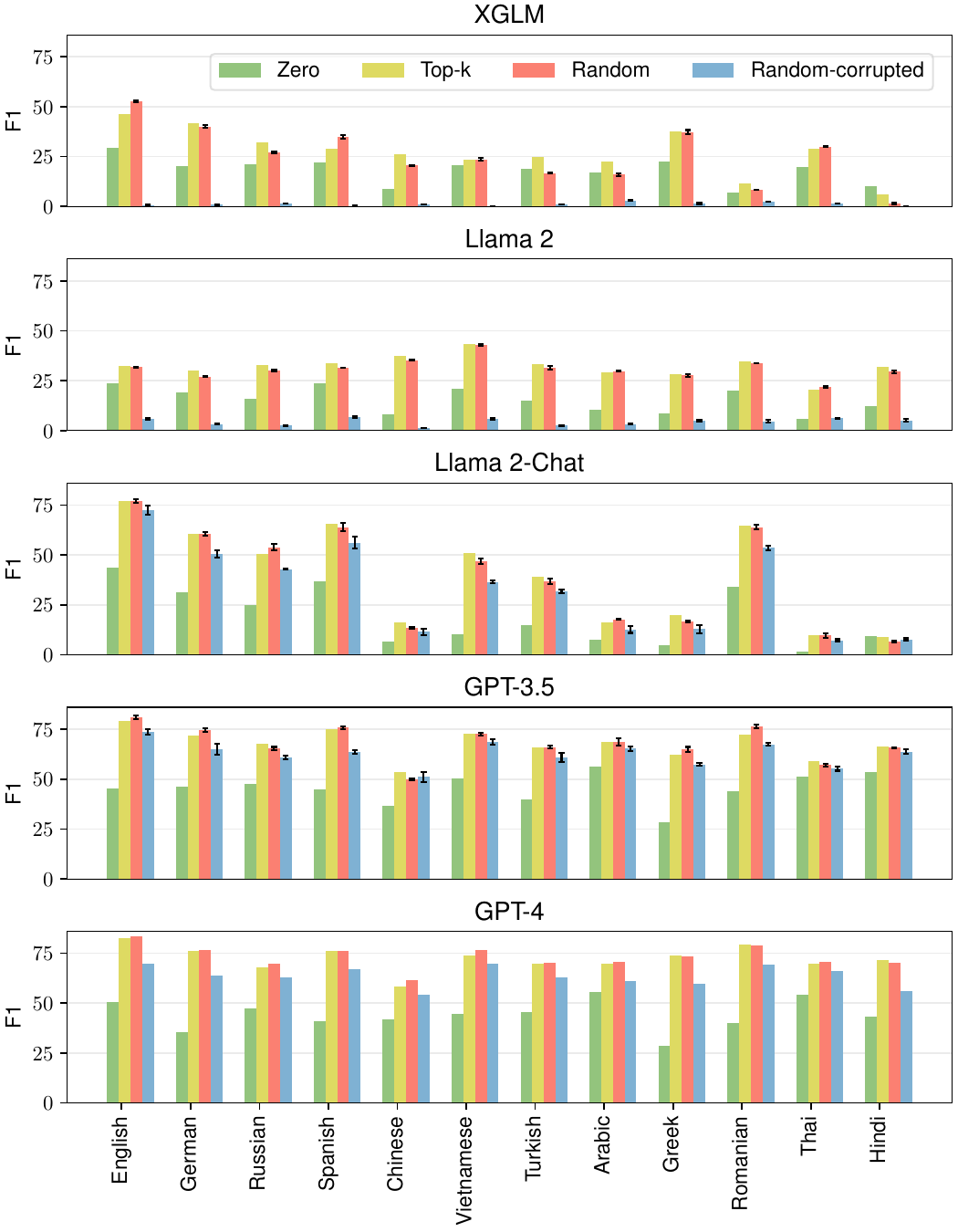}
    \caption{XQuAD}
    \label{fig:xquad_demo}
    \end{subfigure}
    \begin{subfigure}[]{0.49\textwidth}
    \centering
    \includegraphics[width=\textwidth]{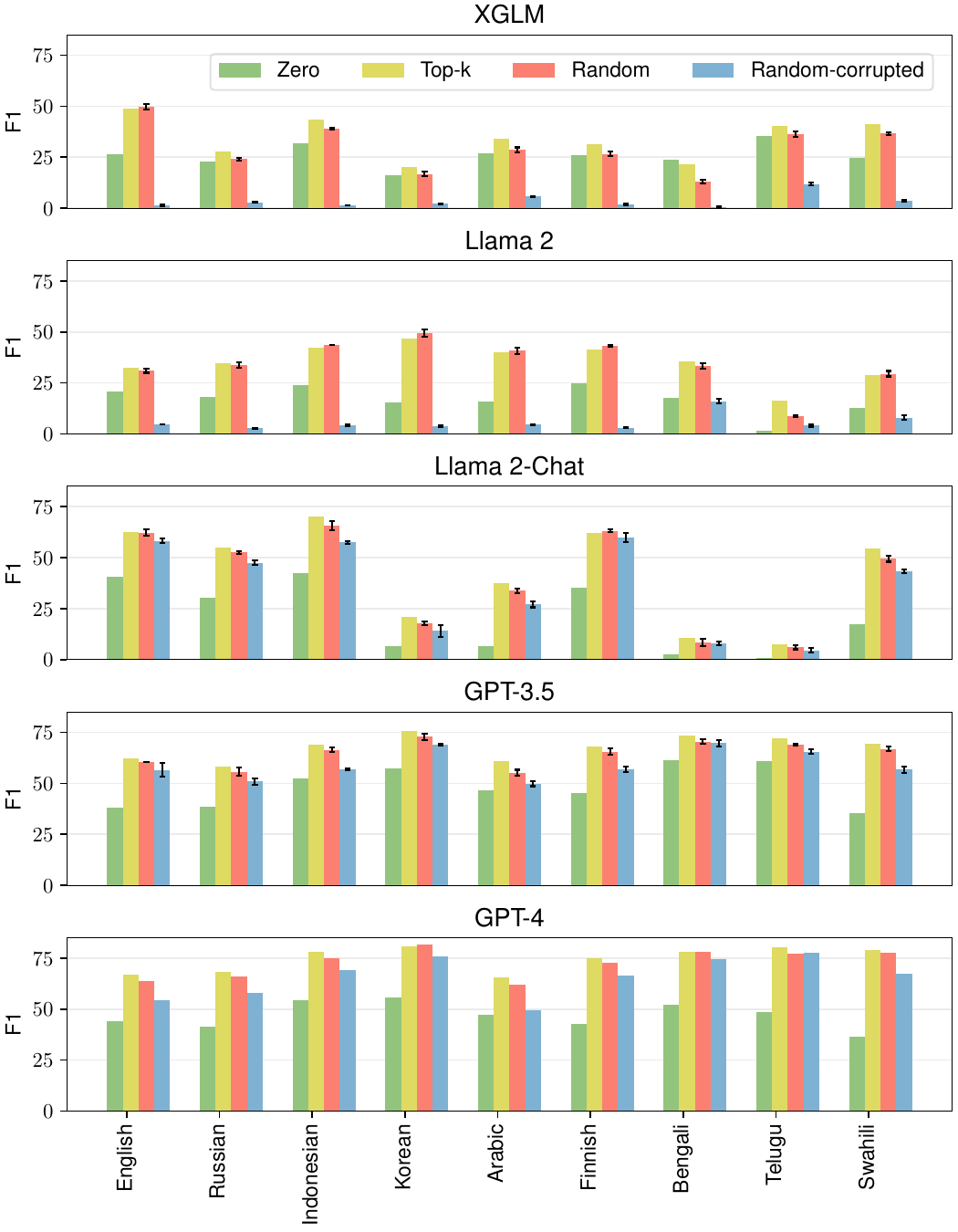}
    \caption{TyDiQA}
    \label{fig:tydiqa_demo}
    \end{subfigure}

\end{figure}

\begin{figure}[h]
    \centering
    \ContinuedFloat
    
    \begin{subfigure}[]{0.49\textwidth}
    \centering
    \includegraphics[width=\textwidth]{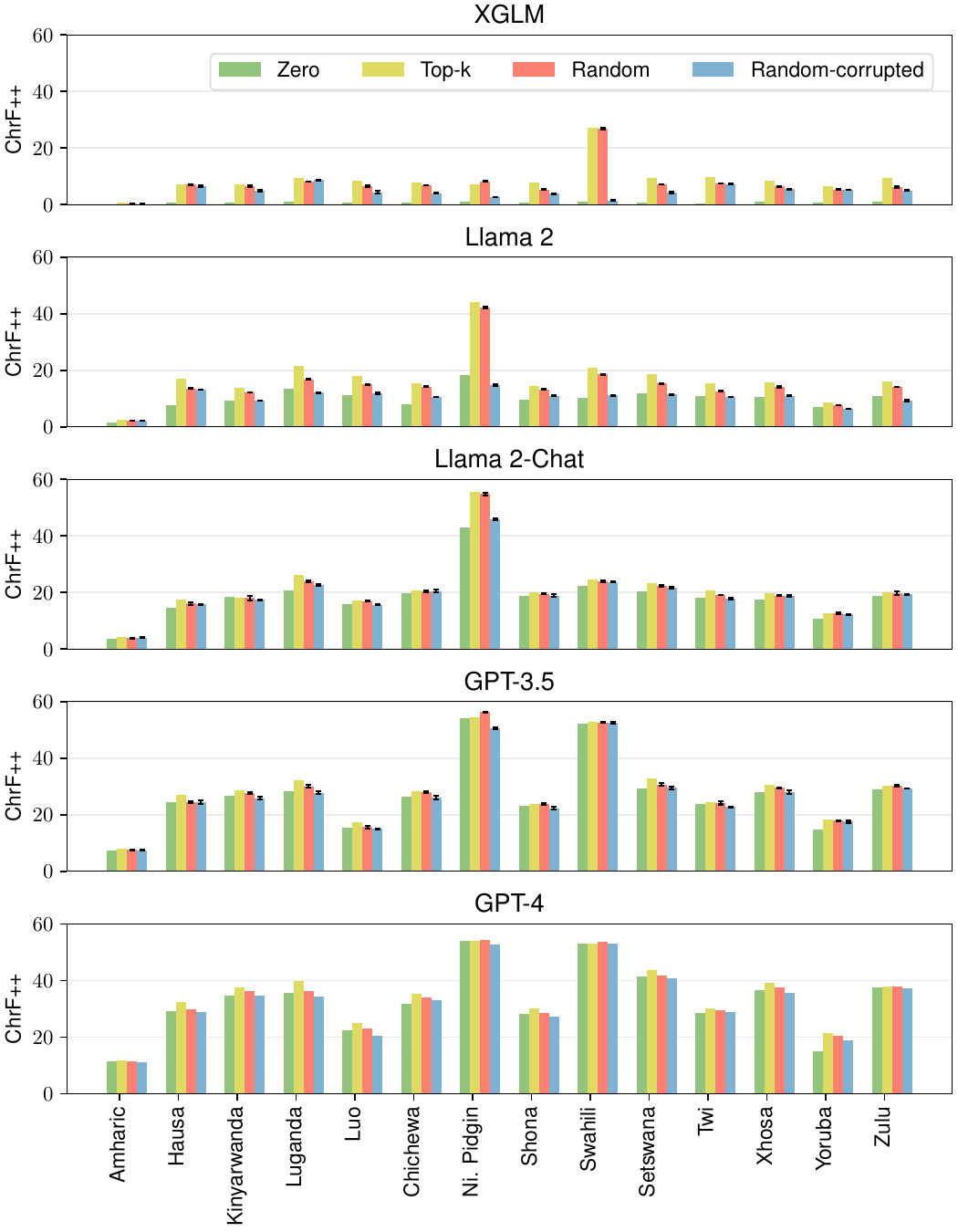}
    \caption{MAFAND (en-xx)}
    \label{fig:mafand_e2t_demo}
    \end{subfigure}
    \begin{subfigure}[]{0.49\textwidth}
    \centering
    \includegraphics[width=\textwidth]{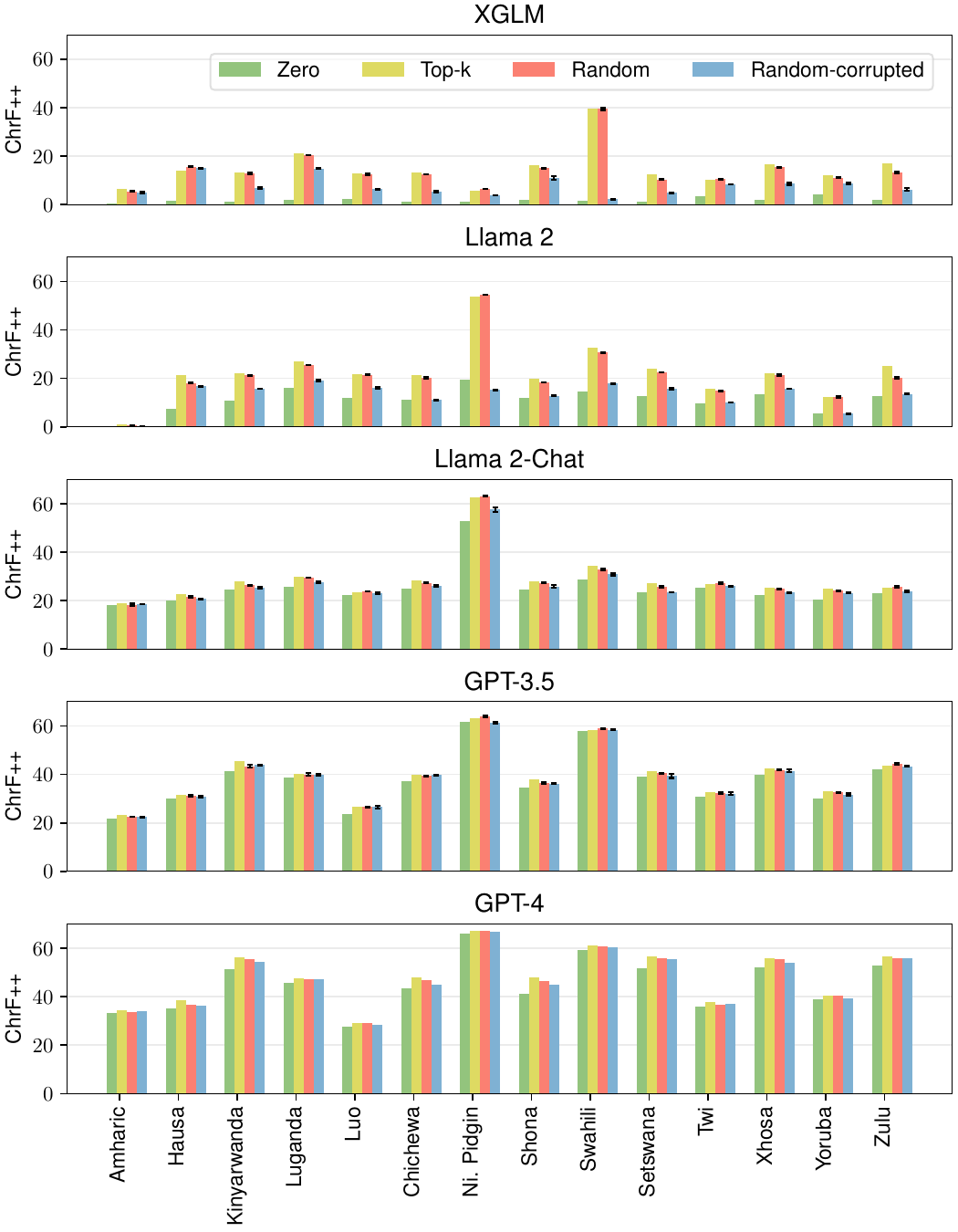}
    \caption{MAFAND (xx-en)}
    \label{fig:mafand_t2e_demo}
    \end{subfigure}
    
\caption{Language-specific performance of 4-shot ICL using different types of demonstrations. We average and report standard deviations over 3 seeds for all models except GPT-4.}
\label{fig:all_demo}
\end{figure}

\definecolor{Color}{RGB}{250, 210, 210}
\definecolor{ColorText}{RGB}{220, 90, 90}

\begin{table*}[]
  \centering
  \resizebox{\linewidth}{!}{
    \begin{tabular}{llllllllllll}
      \toprule
       \textbf{Model} & \textbf{Demonstration} &  \textbf{XNLI} & \textbf{IndicXNLI} & \textbf{PAWS-X} & \textbf{XCOPA} & \textbf{XStoryCloze} & \textbf{AfriSenti} & \textbf{XQuAD} & \textbf{TyDiQA}  & \textbf{MT (en-xx)} & \textbf{MT (xx-en)}  \\ 
        \midrule
       \multirow{4}{*}{XGLM} & \textsc{Zero-shot} & 45.87 & 38.27 & \textbf{54.79} & 57.51 & \textbf{65.19} & 32.71 & 18.16 & 26.01 & 0.79 & 1.89 \\
                      & \textsc{Top-k} & \textbf{45.99} & \textbf{38.85} & 51.72 & \textbf{58.76} & 63.99 & \textbf{44.30} & \textbf{27.54} & \textbf{34.32} & \textbf{9.08} & \textbf{15.05} \\
                      & \textsc{Random} & 41.40$_{0.50}$   & 36.36$_{0.33}$  & 51.48$_{0.33}$ & 58.73$_{0.43}$ & 63.02$_{0.09}$    & 38.68$_{0.39}$  & 25.77$_{0.06}$ & 30.11$_{0.36}$ & 7.77$_{0.09}$    & 14.39$_{0.02}$ \\
                      & \textsc{Random-corrupted} & 40.94$_{0.42}$ & 36.41$_{0.35}$ & 51.04$_{0.28}$ & 58.21$_{0.30}$ & 62.40$_{0.21}$ & 34.90$_{0.42}$ & 1.21$_{0.03}$ & 3.47$_{0.07}$ & 4.59$_{0.01}$ & 7.66$_{0.04}$  \\
       \midrule
       \multirow{4}{*}{Llama 2} & \textsc{Zero-shot} & 44.25      & 37.66       & 59.21      & 56.02      & 65.17         & 32.71       & 15.33      & 16.81      & 10.06        & 11.27 \\
                  & \textsc{Top-k} & \textbf{47.10}       & \textbf{40.15}       & \textbf{59.35}      & 57.69      & \textbf{66.16}         & \textbf{47.25}       & \textbf{32.37}      & \textbf{35.36}      & \textbf{17.29}       & \textbf{22.92} \\
                  & \textsc{Random} & 40.49$_{0.35}$ & 35.98$_{0.24}$ & 57.00$_{0.29}$ & \textbf{57.80$_{0.32}$} & 65.83$_{0.08}$ & 43.08$_{0.02}$ & 31.05$_{0.28}$ & 34.82$_{0.21}$ & 15.14$_{0.01}$ & 21.57$_{0.02}$ \\
                  & \textsc{Random-corrupted} & 39.53$_{0.20}$ & 35.55$_{0.44}$ & 55.85$_{0.92}$ & 57.19$_{0.06}$ & 64.71$_{0.25}$ & 40.81$_{0.36}$ & 4.36$_{0.25}$ & 5.62$_{0.27}$ & 10.35$_{0.04}$ & 13.23$_{0.04}$  \\
       \midrule
      \multirow{4}{*}{Llama 2-Chat} & \textsc{Zero-shot} & 36.10       & 32.32       & \textbf{64.64}      & 44.55      & 57.77         & 31.18       & 18.82      & 20.33      & 18.83        & 25.46 \\
                  & \textsc{Top-k} & 47.53      & 35.73       & 59.36      & \textbf{63.55}      & \textbf{68.82}         & \textbf{45.75}       & \textbf{39.94}      & \textbf{42.38}      & \textbf{21.50}         & \textbf{29.02} \\
                  & \textsc{Random} & 47.81$_{0.85}$ & \textbf{37.09$_{2.57}$} & 61.07$_{1.2}$ & 63.23$_{0.91}$ & 68.39$_{0.14}$ & 43.58$_{0.23}$ & 38.92$_{0.09}$ & 39.96$_{0.30}$ & 20.76$_{0.23}$ & 28.36$_{0.12}$ \\
                  & \textsc{Random-corrupted} & \textbf{48.15$_{1.22}$} & 37.05$_{3.09}$ & 59.59$_{1.13}$ & 63.20$_{0.33}$ & 68.62$_{0.93}$ & 42.81$_{0.11}$ & 32.98$_{0.39}$ & 35.59$_{0.08}$ & 19.63$_{0.04}$ & 26.82$_{0.11}$ \\
       \midrule
     \multirow{4}{*}{GPT-3.5} & \textsc{Zero-shot} & 63.23      & 48.23       & 66.57      & 73.50       & 84.55         & \textbf{53.32}       & 45.25      & 48.52      & 27.39        & 37.77 \\
              & \textsc{Top-k} & \textbf{63.27}      & \textbf{50.45}       & 69.29      & \textbf{80.77}      & 85.23         & 51.86       & 67.82      & \textbf{67.76}      & \textbf{29.20}         & \textbf{39.99} \\
              & \textsc{Random} & 63.09$_{0.88}$ & 49.74$_{1.17}$ & \textbf{71.36$_{0.75}$} & 79.91$_{0.75}$ & \textbf{85.84$_{0.30}$} & 52.52$_{0.21}$ & \textbf{68.16$_{0.36}$} & 64.78$_{0.47}$ & 28.48$_{0.01}$ & 39.56$_{0.03}$ \\
              & \textsc{Random-corrupted} & 62.70$_{1.05}$ & 48.73$_{0.51}$ & 70.71$_{0.66}$ & 79.65$_{0.74}$ & 85.26$_{0.07}$ & 53.14$_{0.47}$ & 62.70$_{0.19}$ & 59.17$_{0.27}$ & 27.09$_{0.18}$ & 39.08$_{0.08}$  \\
       \midrule
     \multirow{4}{*}{GPT-4} & \textsc{Zero-shot} & 70.30       & 65.41       & 74.50       & 88.82      & 96.05         & 58.46       & 44.03      & 46.97      & 32.73        & 45.28 \\
              & \textsc{Top-k} & 76.53      & 67.45       & \textbf{76.14}      & 91.23      & \textbf{96.73}         & 61.68       & 72.44      & \textbf{74.65}      & \textbf{35.06}        & \textbf{48.34} \\
              & \textsc{Random} & 75.77  & 67.64   & 76.07  & \textbf{91.59}  & 96.68     & \textbf{62.36}   & \textbf{73.21}  & 72.77  & 33.85    & 47.69 \\
              & \textsc{Random-corrupted} & \textbf{76.63}  & \textbf{67.68}   & 75.50   & 90.73  & 95.55     & 61.46   & 63.61 & 65.80   & 32.61    & 47.05  \\
      \bottomrule
    \end{tabular}}
  \caption{Performance of different types of demonstrations. For \textsc{random} and \textsc{random-corrupted}, we report the mean and standard deviation across 3 seeds except for GPT-4. Best results for each model and dataset are boldfaced.}
  \label{tab:demos}
\end{table*}

\section{The interplay between demonstrations and templates}
\label{sec:appendix_template}
In this section, we provide supplemental results for Section~\ref{sec:prompt}.

\subsection{Results for individual languages}
\label{sec:appendix_template_languages}
We examine the effect of templates and show language-specific results for XCOPA, AfriSenti, XQuAD and TyDiQA in Figure~\ref{fig:all_template}. In a few cases, we found that formatting-focused templates lead to a decline in performance compared to original templates (e.g., Igbo and Mozambican Portuguese in AfriSenti with GPT-3.5). This can be attributed to the model's sensitivity to prompts, highlighting the potential of automatic prompt engineering. Still, formatting-focused template can largely narrow the performance gap between 0-shot and 4-shot in a broad context. 

\begin{figure*}[h]
    \begin{subfigure}[]{\linewidth}
    \centering
    \includegraphics[width=\linewidth]{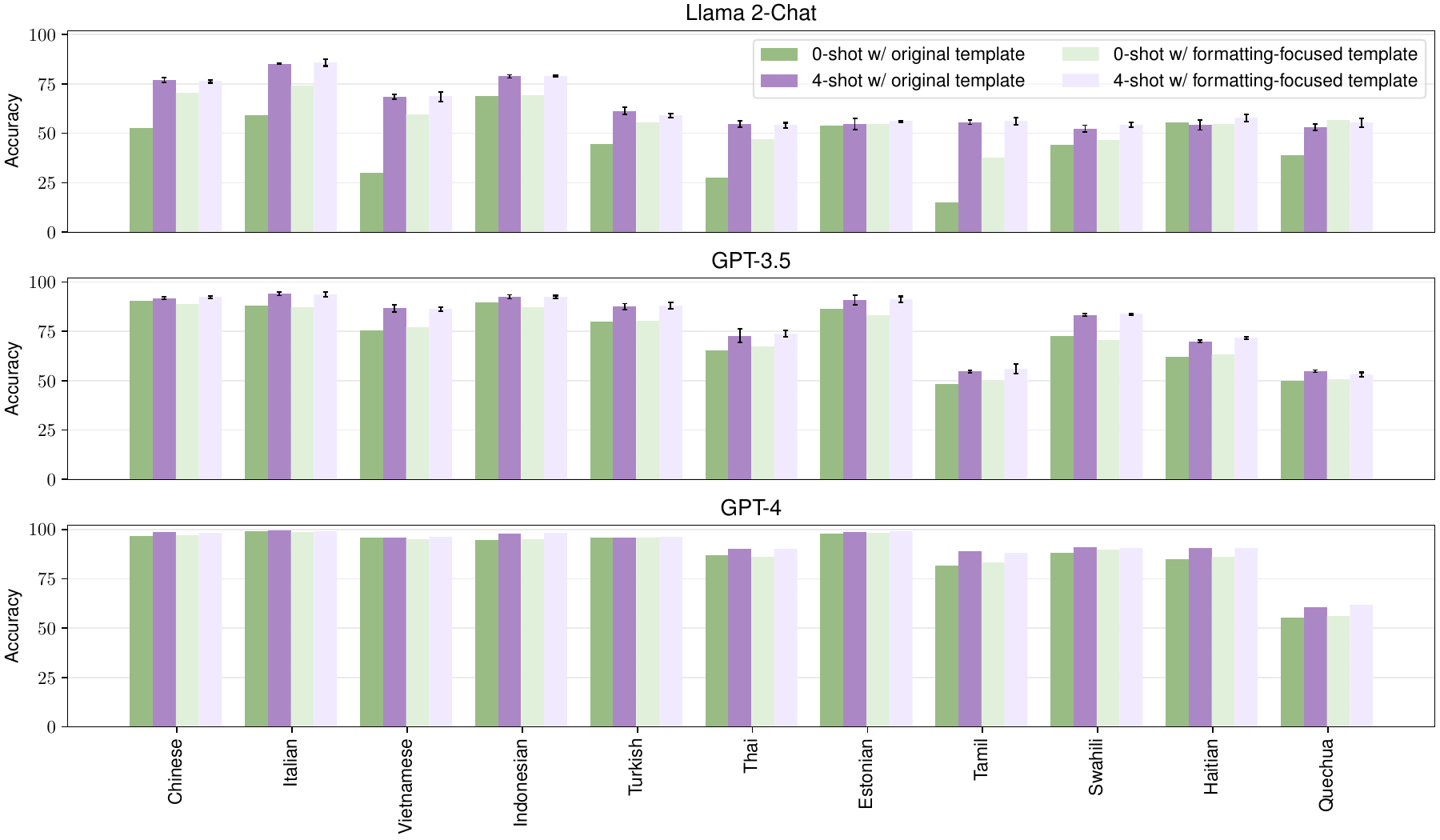}
    \caption{XCOPA}
    \label{fig:xcopa_template}
    \end{subfigure}

    \vspace{1em}

    \begin{subfigure}[]{\linewidth}
    \centering
    \includegraphics[width=\linewidth]{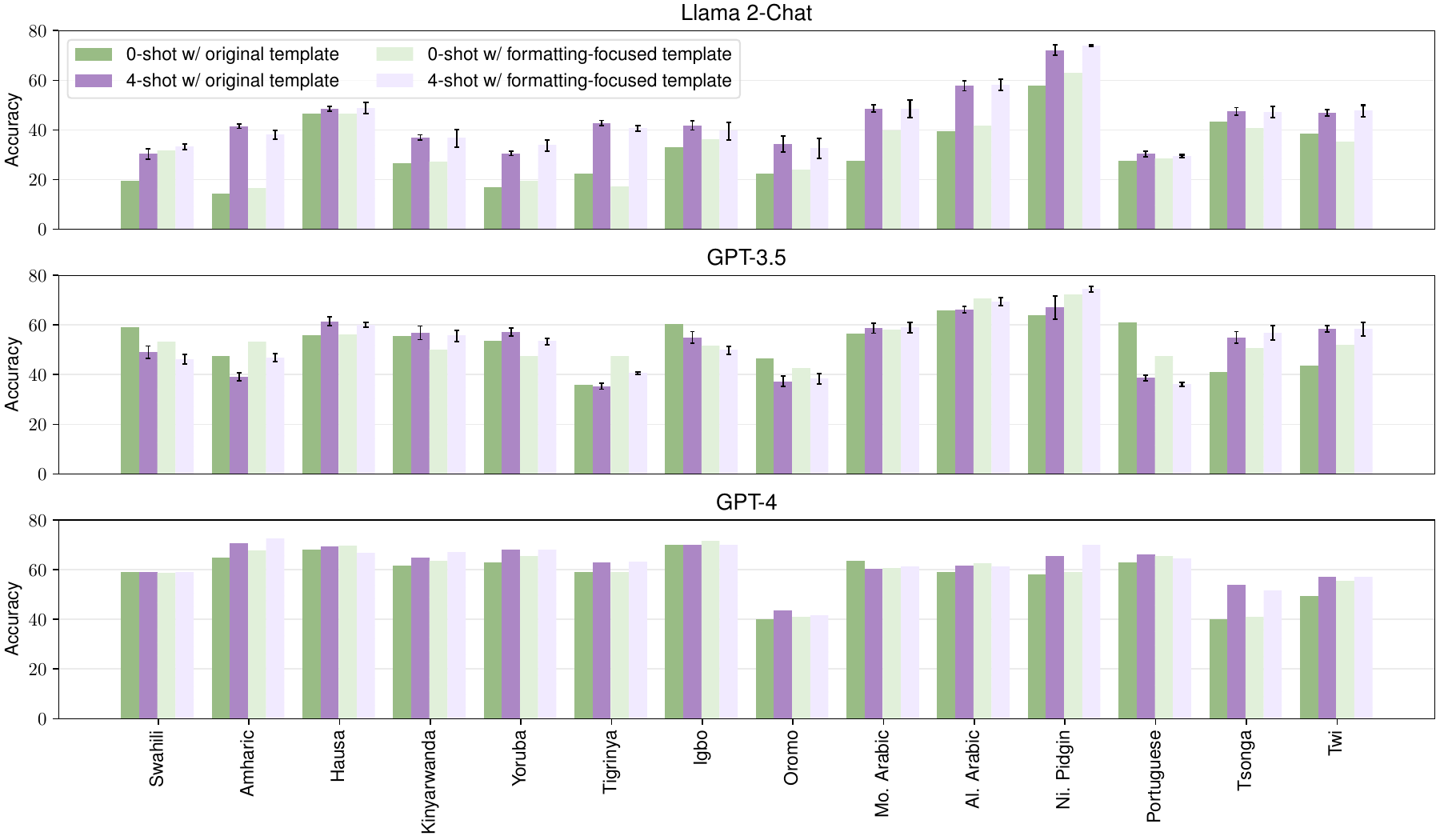}
    \caption{AfriSenti}
    \label{fig:afrisenti_template}
    \end{subfigure}
\end{figure*}

\begin{figure*}[h]
    \ContinuedFloat
    \begin{subfigure}[]{\linewidth}
    \centering
    \includegraphics[width=\linewidth]{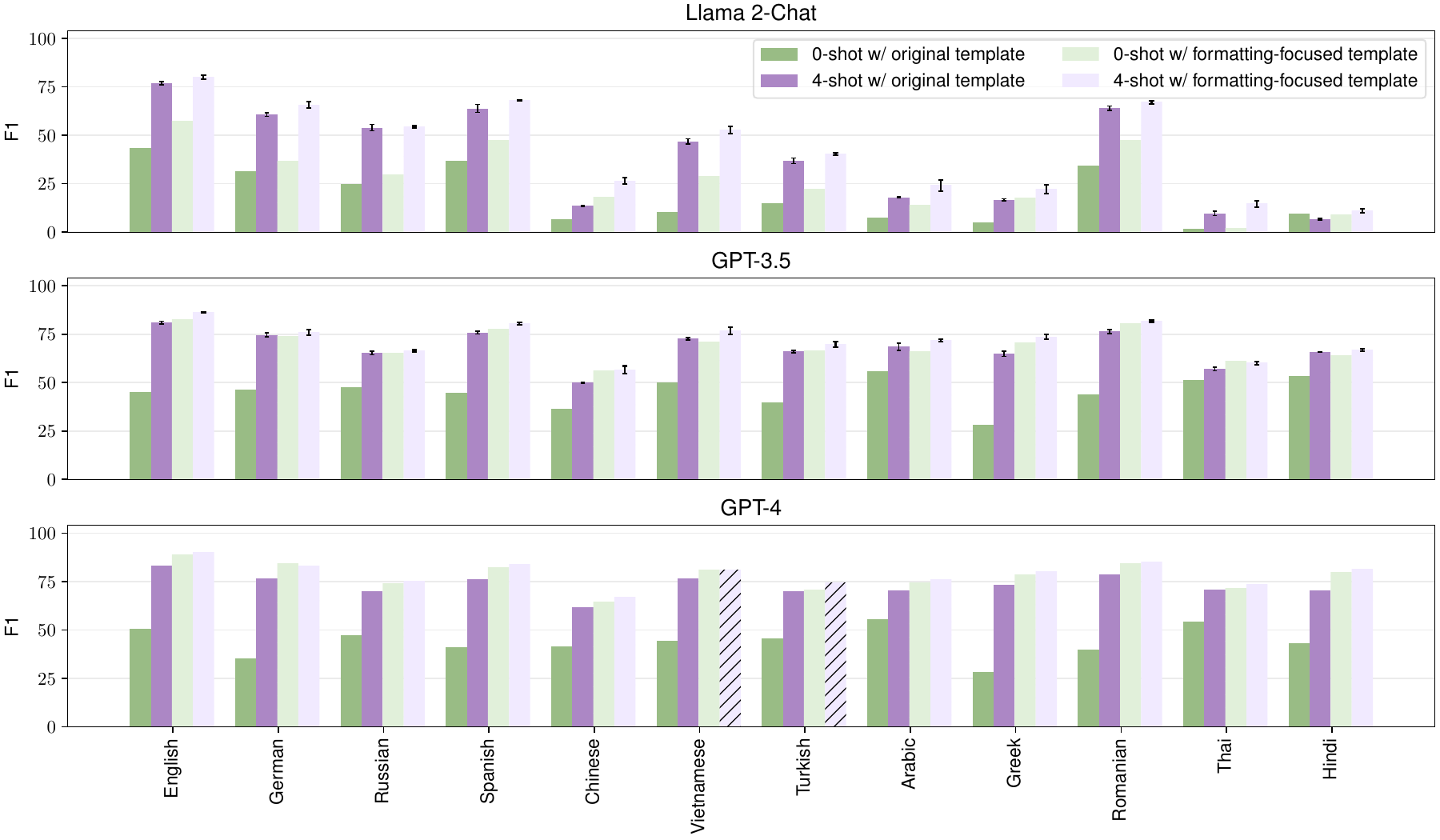}
    \caption{XQuAD}
    \label{fig:xquad_template}
    \end{subfigure}

     \vspace{1em}
    
    \begin{subfigure}[]{\linewidth}
    \centering
    \includegraphics[width=\linewidth]{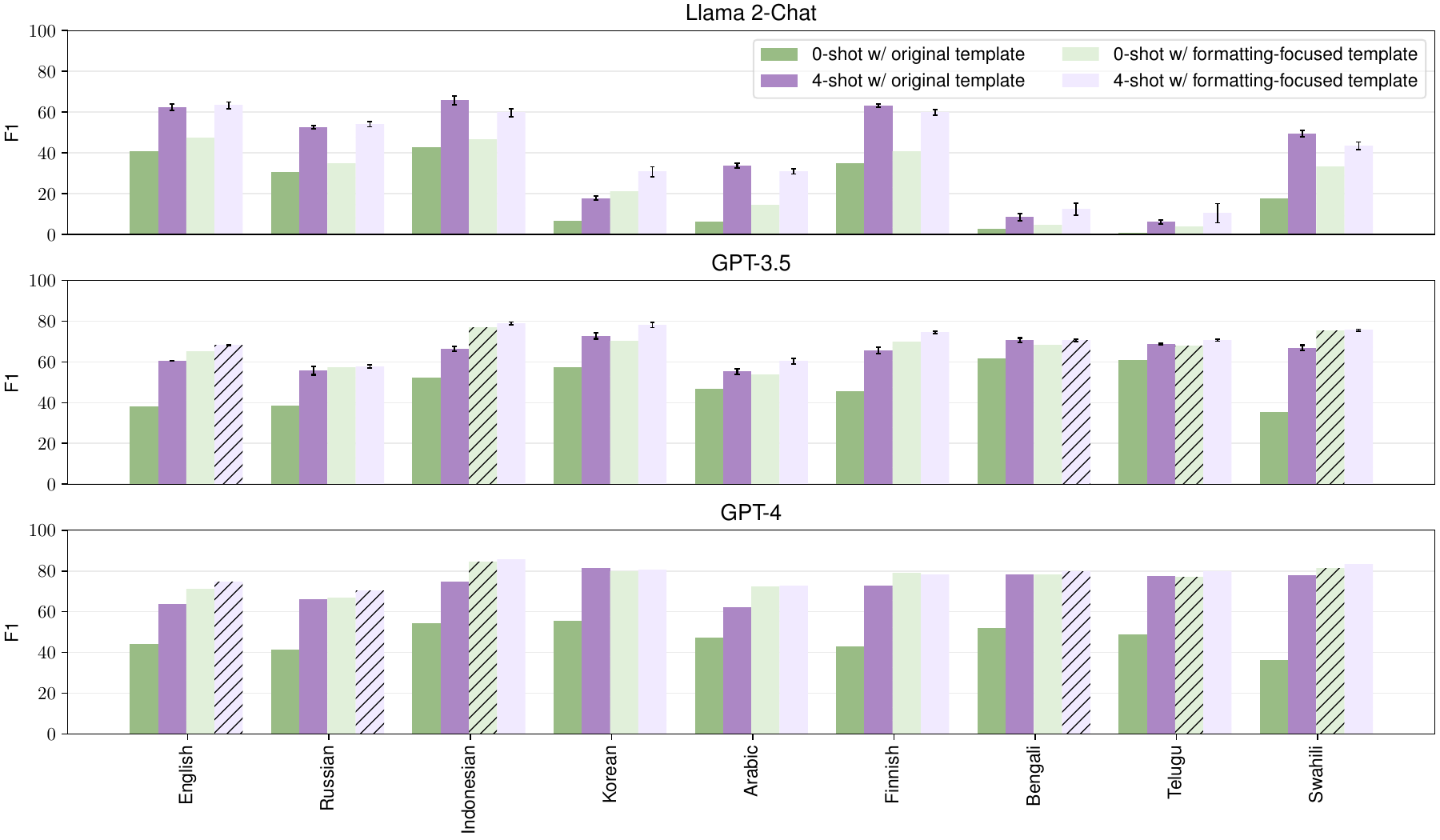}
    \caption{TyDiQA}
    \label{fig:tydiqa_template}
    \end{subfigure}

    \caption{Effect of using different templates on 0-shot and 4-shot performance for XCOPA, AfriSenti, and TyDiQA. Few-shot results are averaged across 3 seeds except for GPT-4.}
    \label{fig:all_template}
\end{figure*}

\newcommand{\slashn}{\textbackslash n}
\definecolor{LightCyan}{rgb}{0.88,1,1}
\definecolor{TextColor}{rgb}{0.9,0.2,0.2}

\begingroup
\renewcommand{\arraystretch}{1.2} %

\begin{table*}[h!]
\centering
\resizebox{\linewidth}{!}{
    \begin{tabular}{lll}
    \toprule
    \textbf{Task}  & \textbf{Pattern}    & \textbf{Verbalizer}   \\ 
    \midrule
    \rowcolor{LightCyan}NLI 
             & \texttt{\{premise\}}, right? \texttt{\{label\}}, \texttt{\{hypothesis\}} 
             &  Yes || Also || No \\ 
    PAWS-X
            &\texttt{\{sentence1\}}, right? \texttt{\{label\}}, \texttt{\{sentence2\}} 
            &  No || Yes   \\ 
    \rowcolor{LightCyan}XCOPA                  
            & \texttt{\{premise\}} \{\% if question == “cause" \%\}because\{\% else \%\}          &  \texttt{\{choice1\}} || \texttt{\{choice2\}}     \\
            \rowcolor{LightCyan}&so\{\% endif \%\} \texttt{\{label\}} & \\
    XStoryCloze   
            & \texttt{\{input\_sentence\_1\}} \texttt{\{input\_sentence\_2\}}
            & \texttt{\{sentence\_quiz\_1\}} || \\
            &\texttt{\{input\_sentence\_3\}} \texttt{\{input\_sentence\_4\}}  \texttt{\{label\}} 
            &\texttt{\{sentence\_quiz\_2\}} \\
    \rowcolor{LightCyan}AfriSenti     
            & \texttt{\{tweet\}} The sentiment of the previous sentence is \texttt{\{label\}}       &  positive || neutral || negative \\ 
    \rowcolor{LightCyan}QA 
            & \texttt{\{context\}}\slashn Q:\texttt{\{question\}}\slashn A:\texttt{\{answer\}} 
            & \texttt{\{answer\}}  \\
    MT 
            & \texttt{\{source\_sentence\}} = \texttt{\{target\_sentence\}} & \texttt{\{target\_sentence\}} \\ 
    \bottomrule
    \end{tabular}}
\caption{Prompting templates for XGLM and Llama 2 following \citet{brown2020language} and \citet{lin-etal-2022-shot}.}
\label{tab:template_xglm}
\end{table*}

\definecolor{LightCyan}{rgb}{0.88,1,1}
\definecolor{TextColor}{rgb}{0.9,0.2,0.2}

\begingroup
\renewcommand{\arraystretch}{1.2} %

\begin{table*}[h!]
\centering
\resizebox{\linewidth}{!}{
    \begin{tabular}{lll}
    \toprule
    \textbf{Task}  & \textbf{Pattern}    & \textbf{Verbalizer}   \\ 
    \midrule
    \rowcolor{LightCyan}
    NLI
             & \texttt{\{premise\}} Based on the previous passage, is it true that 
             &  Yes || Maybe || No  \\
    \rowcolor{LightCyan} 
             &  \texttt{\{hypothesis\}}? Yes, No, or Maybe? \texttt{\{label\}} &  \\ 
    PAWS-X
            & Sentence 1: \texttt{\{sentence1\}}\slashn
            &  No || Yes  \\
            & Sentence 2: \texttt{\{sentence2\}}\slashn
            & \\
            & Question: Can we rewrite Sentence 1 to Sentence 2? Yes or No? &\\
            & \texttt{\{label\}} &\\
    \rowcolor{LightCyan}XCOPA                  
            & \texttt{\{premise\}} \{\% if question == “cause" \%\}This happened because...&\\
    \rowcolor{LightCyan}
            &\{\% else \%\} As a consequence...\{\% endif \%\}\slashn 
            & \\
    \rowcolor{LightCyan}
            &Help me pick the more plausible option:\slashn 
            & \texttt{\{choice1\}} || \texttt{\{choice2\}} \\
    \rowcolor{LightCyan}
            & - \texttt{\{choice1\}}\slashn 
            & \\
    \rowcolor{LightCyan}
            & - \texttt{\{choice2\}}\slashn 
            & \\ 
    \rowcolor{LightCyan}
            & \texttt{\{label\}} & \\
    XStoryCloze   
            & \texttt{\{input\_sentence\_1\}} \texttt{\{input\_sentence\_2\}} 
            & \\
            &\texttt{\{input\_sentence\_3\}} \texttt{\{input\_sentence\_4\}}\slashn 
            &\\
            & What is a possible continuation for the story given the following
            & \texttt{\{sentence\_quiz\_1\}} || \\
            & options?\slashn 
            &\texttt{\{sentence\_quiz\_2\}} \\
            & - \texttt{\{sentence\_quiz\_1\}}\slashn 
            & \\
           & - \texttt{\{sentence\_quiz\_2\}}\slashn &\\ 
           &\texttt{\{label\}} 
           & \\
    \rowcolor{LightCyan}
    AfriSenti     
            & \texttt{\{tweet\}} Would you rate the previous sentence as positive, 
            & positive || neutral || negative \\
    \rowcolor{LightCyan}
            &  neutral or negative? \texttt{\{label\}} & \\ 
    QA 
           & \texttt{\{context\}}\slashn Q:\texttt{\{question\}}\slashn Referring to the passage above,  
           & \texttt{\{answer\}}  \\
           & the correct answer to the given question is:\texttt{\{answer\}} 
           & \\
    \rowcolor{LightCyan}
    MT
          & Translate the following \texttt{\{src\_language\}} text to \texttt{\{tgt\_language\}}:\slashn & \texttt{\{tgt\_sentence\}}\\
          \rowcolor{LightCyan}
          & \texttt{\{src\_sentence\}}\slashn \texttt{\{tgt\_sentence\}} & \\
    \bottomrule
    \end{tabular}}
\caption{Prompting templates for BLOOMZ and mT0 following \citet{muennighoff-etal-2023-crosslingual} and \citet{bach-etal-2022-promptsource}.}
\label{tab:template_bloomz}
\end{table*}

\definecolor{LightCyan}{rgb}{0.88,1,1}
\definecolor{TextColor}{rgb}{0.9,0.2,0.2}

\begingroup
\renewcommand{\arraystretch}{1.2} %

\begin{table*}[h!]
\centering
\resizebox{\linewidth}{!}{
    \begin{tabular}{ll}
    \toprule
    \textbf{Task}  & \textbf{Template}  \\ 
    \midrule
    NLI      
             &  \textbf{task instruction}: You are an NLP assistant whose purpose is to solve Natural Language Inference \\
             & (NLI) problems in <EVALUATION\_LANGUAGE>.     NLI is the task of determining the inference relation \\
             & between two (short, ordered) texts: entailment, contradiction, or neutral. Answer as concisely as \\
             & possible in the same format as the examples below: \\
             &\textbf{pattern}: \texttt{\{premise\}}\slashn Question: \texttt{\{hypothesis\}}\slashn True, False, or Neither? \\
             &\textbf{verbalizer}: True || Neither || False  \\
    \midrule
    PAWS-X  
             &  \textbf{task instruction}: You are an NLP assistant whose purpose is to perform Paraphrase Identification in  \\
             & <EVALUATION\_LANGUAGE>. The goal of Paraphrase Identification is to determine whether a pair\\
             & of sentences have the same meaning. Answer as concisely as possible in the same format as the \\
             & examples below: \\
             &\textbf{pattern}: \texttt{\{sentence1\}}\slashn Question: \texttt{\{sentence2\}}\slashn True or False? \\
             &\textbf{verbalizer}: False || True  \\
    \midrule
    XCOPA 
             &  \textbf{task instruction}: You are an NLP assistant whose purpose is to perform  open-domain commonsense   \\
             & causal reasoning in <EVALUATION\_LANGUAGE>.  You will be provided a premise and two alternatives, \\
             & where the task is to select the alternative    that more plausibly has a causal relation with the premise.  \\
             & Answer as concisely as possible in the same format as the examples below:  \\
             &\textbf{pattern}: \\
             &Premise: \texttt{\{premise\}}\slashn What is the \texttt{\{question\}}? Pick the more plausible option:\slashn \\
             &1: \texttt{\{choice1\}}\slashn 2: \texttt{\{choice2\}}\slashn \\
             &You should tell me the choice number in this format 'Choice number:' \\
             &\textbf{verbalizer}: Choice number: 1 || Choice number: 2  \\
    \midrule
    XStoryCloze 
             & \textbf{task instruction}: You are an NLP assistant whose purpose is to perform  open-domain commonsense   \\
             & causal reasoning in <EVALUATION\_LANGUAGE>.  You will be provided a four-sentence story and two \\
             & continuations, where the task is to select the correct ending. Answer as concisely as possible in the same \\
             &  format as the examples below:  \\
             &\textbf{pattern}: \\
             &Story: \texttt{\{input\_sentence\_1\}} \texttt{\{input\_sentence\_2\}} \texttt{\{input\_sentence\_3\}} \texttt{\{input\_sentence\_4\}}\slashn \\
             &What is a possible continuation for the story? Pick the more plausible option:\slashn \\
             &1: \texttt{\{sentence\_quiz1\}}\slashn 2: \texttt{\{sentence\_quiz2\}}\slashn \\
             &You should tell me the choice number in this format 'Choice number:' \\
             &\textbf{verbalizer}: Choice number: 1 || Choice number: 2  \\
    \midrule
    AfriSenti  
             &  \textbf{task instruction}: You are an NLP assistant whose purpose is to perform Sentiment Analysis in  \\
             & <EVALUATION\_LANGUAGE>. Sentiment Analysis is the task of determining the sentiment, \\
             & opinion or emotion expressed in a textual data. Give your answer as a single word, "positive", "neutral"\\
             &or "negative". \\
             &\textbf{pattern}: Does this statement 
             ``\texttt{\{tweet\}}'' have a \{positive neutral or negative\} sentiment? Labels only \\
             &\textbf{verbalizer}: positive || neutral || negative  \\
    \midrule
    QA 
             & \textbf{task instruction}: You are an NLP assistant whose purpose is to solve reading comprehension  \\
             & problems in <EVALUATION\_LANGUAGE>. You will be provided questions on a set of passages and\\
             &  you will need to provide the answer as it appears in the passage. The answer should be in the same \\
             &  language as the question and the passage. \\
             &\textbf{pattern}: \\
             &\texttt{\{context\}}\slashn Q: \texttt{\{question\}}\slashn Referring to the passage above, the correct answer to the given question is: \\
             &\textbf{verbalizer}: \texttt{\{answer\}}  \\
    \midrule
    MT 
            &\textbf{pattern}: Translate the following \texttt{\{src\_language\}} text to \texttt{\{tgt\_language\}}: \texttt{\{src\_sentence\}} \\
            &\textbf{verbalizer}: \texttt{\{tgt\_sentence\}}  \\
    \bottomrule
    \end{tabular}}
\caption{Prompting templates for chat models following \citet{ahuja-etal-2023-mega} and \citet{ojo2023good}. We add language identifiers in task instructions as it is an effective strategy for improving multilingual prompting~\citep{huang-etal-2023-languages}. }
\label{tab:template_openai}
\end{table*}

\begin{table*}[]
\centering
\resizebox{\linewidth}{!}{
    \begin{tabular}{ll}
    \toprule
    \textbf{Task}  & \textbf{Template}  \\ 
    \midrule
    XCOPA 
             &  \textbf{task instruction}: You are an NLP assistant whose purpose is to perform  open-domain commonsense   \\
             & causal reasoning in <EVALUATION\_LANGUAGE>.  You will be provided a premise and two alternatives, \\
             & where the task is to select the alternative    that more plausibly has a causal relation with the premise.  \\
             & Answer as concisely as possible in the same format as the examples below:  \\
             &\textbf{pattern}: \\
             &Premise: \texttt{\{premise\}}\slashn What is the \texttt{\{question\}}? Pick the more plausible option:\slashn \\
             &1: \texttt{\{choice1\}}\slashn 2: \texttt{\{choice2\}}\slashn \\
             & \textcolor{TextColor}{This is very important: Do not repeat the question and no explanation.} \\
             &You should tell me the choice number in this format 'Choice number:' \\
             &\textbf{verbalizer}: Choice number: 1 || Choice number: 2  \\
    \midrule
    AfriSenti  
             &  \textbf{task instruction}: You are an NLP assistant whose purpose is to perform Sentiment Analysis in  \\
             & <EVALUATION\_LANGUAGE>. Sentiment Analysis is the task of determining the sentiment, \\
             & opinion or emotion expressed in a textual data. Give your answer as a single word, "positive", "neutral"\\
             &or "negative". \\
             &\textbf{pattern}: Does this statement 
             ``\texttt{\{tweet\}}'' have a \{positive neutral or negative\} sentiment? \\
             &\textcolor{TextColor}{This is very important: Do not repeat the question and no explanation.} Labels only \\
             &\textbf{verbalizer}: positive || neutral || negative  \\
    \midrule
    QA 
             & \textbf{task instruction}: You are an NLP assistant whose purpose is to solve reading comprehension  \\
             & problems in <EVALUATION\_LANGUAGE>.     Answer the question from the given passage. \textcolor{TextColor}{Your answer} \\
             & \textcolor{TextColor}{should be directly extracted from the passage and be a single entity, name, or number, not a sentence.} \\
             &\textbf{pattern}: \\
             &\texttt{\{context\}}\slashn Q: \texttt{\{question\}}\slashn \textcolor{TextColor}{This is very important: Your answer should be directly extracted from the} \\
             & \textcolor{TextColor}{passage and be a single entity, name, or number, not a sentence.} \\
             &\textbf{verbalizer}: \texttt{\{answer\}}  \\
    \bottomrule
    \end{tabular}}
    \caption{Formatting-focused templates for chat models. We augmented the original templates in Table~\ref{tab:template_openai} with \textcolor{TextColor}{formatting-focused instructions}.}
\label{tab:template_openai_1}
\end{table*}

\endgroup

\end{document}